\documentclass[a4paper,twoside]{article}

\usepackage{epsfig}
\usepackage{subcaption}
\usepackage{calc}
\usepackage{amssymb}
\usepackage{amstext}
\usepackage{amsmath}
\usepackage{amsthm}
\usepackage{multicol}
\usepackage{pslatex}
\usepackage{apalike}
\usepackage{array}
\usepackage{balance}
\usepackage{csquotes}

\usepackage{booktabs}
\usepackage{multirow}
\usepackage{hyperref}
\usepackage{SCITEPRESS}     

\newcolumntype{L}[1]{>{\raggedright\let\newline\\\arraybackslash\hspace{0pt}}m{#1}}
\newcolumntype{C}[1]{>{\centering\let\newline\\\arraybackslash\hspace{0pt}}m{#1}}
\newcolumntype{R}[1]{>{\raggedleft\let\newline\\\arraybackslash\hspace{0pt}}m{#1}}

\newcommand{\etal}{\emph{et al.}~}
\newcommand{\eg}{\emph{e.g.}~}
\newcommand{\ie}{\emph{i.e.}~}

\begin{document}

\title{Learn to See by Events: \\ Color Frame Synthesis from Event and RGB Cameras}

\author{\authorname{Stefano Pini\sup{1}\orcidAuthor{0000-0002-9821-2014}, Guido Borghi \sup{2}\orcidAuthor{0000-0003-2441-7524} and Roberto Vezzani\sup{1,2}\orcidAuthor{0000-0002-1046-6870}}
\affiliation{\sup{1}DIEF - Dipartimento di Ingegneria ``Enzo Ferrari'', University of Modena and Reggio Emilia, Italy}
\affiliation{\sup{2}AIRI - Artificial Intelligence Research and Innovation Center, University of Modena and Reggio Emilia, Italy}
\email{\{s.pini, guido.borghi, roberto.vezzani\}@unimore.it}}

\keywords{Event Cameras, Event Frames, Simulated Event Frames, Color Frame Synthesis, Automotive}

\abstract{Event cameras are biologically-inspired sensors that gather the temporal evolution of the scene. They capture pixel-wise brightness variations and output a corresponding stream of asynchronous events.
Despite having multiple advantages with respect to traditional cameras, their use is partially prevented by the limited applicability of traditional data processing and vision algorithms.
To this aim, we present a framework which exploits the output stream of event cameras to synthesize RGB frames, relying on an initial or a periodic set of color key-frames and the sequence of intermediate events.
Differently from existing work, we propose a deep learning-based frame synthesis method, consisting of an adversarial architecture combined with a recurrent module. 
Qualitative results and quantitative per-pixel, perceptual, and semantic evaluation on four public datasets confirm the quality of the synthesized images.
}

\onecolumn
\maketitle
\normalsize
\setcounter{footnote}{0}
\vfill

\section{\uppercase{Introduction}} \label{sec:introduction}
\begin{figure*}[h!] 
\begin{center}
  \includegraphics[width=1\linewidth]{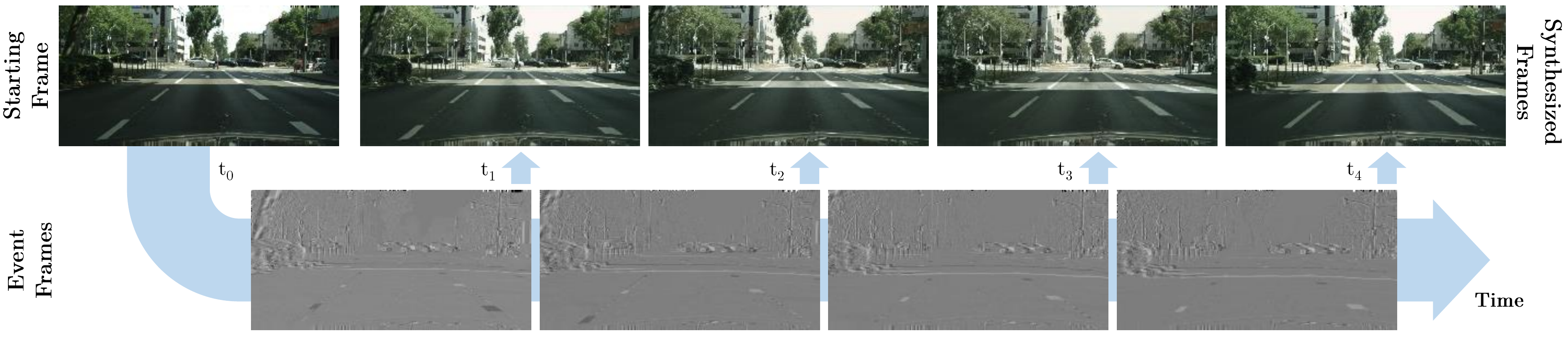}
\end{center}
  \caption{Sample frames synthesized by the proposed framework. Given an initial RGB frame at time $t_0$ and a set of following event frames at time $t_1, ..., t_n$ as input, the proposed framework accordingly synthesizes an RGB frame for each time step.}
\label{fig:front}
\end{figure*}

Event cameras are neuromorphic optical sensors capable of asynchronously capturing pixel-wise brightness variations, \textit{i.e.}~\textit{events}. They are gaining more and more attention from the computer vision community thanks to their extremely high temporal resolution, low power consumption, reduced data rate, and high dynamic range~\cite{gallego2018unifying}.\\
Moreover, event cameras filter out redundant information as their output intrinsically embodies only the temporal dynamics of the recorded scene, ignoring static and non-moving areas.
On the other hand, standard intensity cameras with an equivalent frame rate are able to acquire the whole complexity of the scene, including textures and colors. However, they usually require a huge amount of memory to store the collected data, along with a high power consumption and a low dynamic range~\cite{maqueda2018event}.\\
Given the availability of many mature computer vision algorithms for standard images, being able to apply them on event data, without the need of designing specific algorithms or collecting new datasets, could contribute to the spread of event sensors.\\
Differently from existing works, which are built on filter- and optimization-based algorithms~\cite{brandli2014real,munda2018real,scheerlinck2018continuous}, in this paper we investigate the use of deep learning-based approaches to interpolate frames from a low frame rate RGB camera using event data.
\\
In particular, we propose a model that synthesizes color or gray-level frames preserving high-quality textures and details (Fig.~\ref{fig:front}) thus allowing the use of traditional vision algorithms like object detection and semantic segmentation networks. 
\\
We explore the use of a conditional adversarial network~\cite{mirza2014conditional} in conjunction with a recurrent module to estimate RGB frames, relying on an initial or a periodic set of color key-frames and a sequence of event frames, \textit{i.e.}~frames that collect events occurred in a certain amount of time.
\\
Moreover, we propose to use simulated event data, obtained by means of image differences, to train our model: this solution leads to two significant advantages.
First, event-based methods can be evaluated on standard datasets with annotations, which are often not available in the event domain.
Second, learned models can be trained on simulated event data and used with real event data, unseen during the training procedure.
In general, we propose to shift from event-based context to a domain where more expertise is available in terms of mature vision algorithms.\\
As a case study, we embrace the automotive context, in which event and intensity cameras could cover a variety of applications \cite{borghi2019driver,frigieri2017fast}. 
For instance, the growing number of high-quality cameras placed on recent cars implies the use of a large bandwidth in the internal and the external network: sending only key-frames and events might be a way to reduce the bandwidth requirements, still maintaining a high temporal resolution.
\\
We probe the feasibility of the proposed model testing it on four automotive and publicly-released datasets, namely \textit{DDD17}~\cite{binas2017ddd17}, \textit{MVSEC}~\cite{zhu2018multivehicle}, \textit{Kitti}~\cite{Geiger2013IJRR}, and 
\textit{Cityscapes}~\cite{cordts2016cityscapes}.\\
Summarizing, our contributions are threefold:
    \textit{i}) we propose a framework based on a conditional adversarial network that performs the synthesis of \textit{color} or gray-level frames.
    \textit{ii}) we investigate the use of simulated event frames to train systems able to work with real event data;
    \textit{iii}) we probe the effectiveness of the proposed method employing four public automotive datasets, investigating the ability to generate \textit{realistic} images, preserving colors, objects, and the semantic information of the scene.

\section{\uppercase{Related Work}}
Event-based vision has recently attracted the attention of the computer vision community.
In the last years, event-based cameras, also known as neuromorphic sensors or \textit{Dynamic Vision Sensors} (DVSs)~\cite{lichtsteiner2008128}, have been mainly explored for monocular~\cite{rebecq2016emvs} and stereo depth estimation~\cite{andreopoulos2018low,zhou2018semi}, optical flow prediction~\cite{gallego2018unifying} as well as for real time feature detection and tracking \cite{rameshlong,mitrokhin2018event} and ego-motion estimation~\cite{maqueda2018event,gallego2018event}.
Moreover, various classification tasks were addressed employing event-based data, as classification of faces~\cite{lagorce2017hots} and gestures~\cite{lungu2017live}.
\begin{figure*}[t!] 
\begin{center}
  \includegraphics[width=0.99\linewidth]{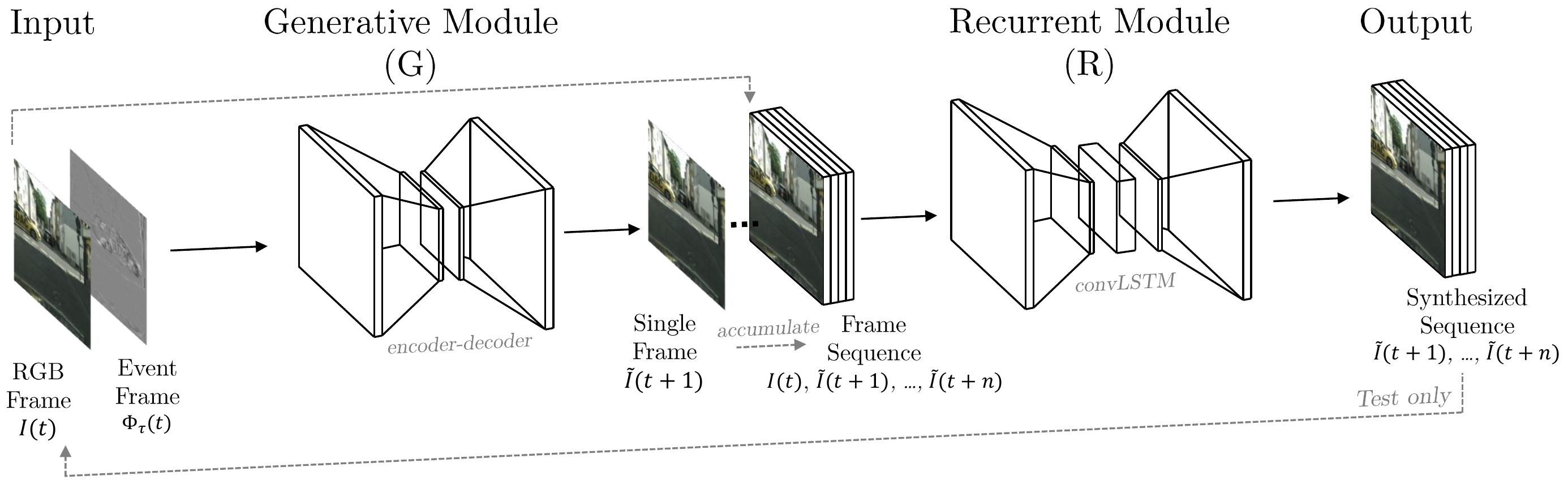}
\end{center}
  \caption{Overview of the proposed framework. The input of the \textit{Generative Module} is the initial intensity frame $I(t)$ and the corresponding event frame $\Phi_{\tau}(t)$. 
  $I(t)$ and the generated frames $\tilde{I}(t_i)$ 
  are then stacked and used as input for the \textit{Recurrent Module} that 
  learns how to refine contextual details and maintain the temporal coherence between consecutive frames. 
  At testing time, the output of the \textit{Recurrent Module} is used in the next forward step as the new input frame of the framework.}
\label{fig:overview}
\end{figure*}

Recently, a limited amount of work focused on the reconstruction of intensity images or videos from event cameras.
Bardow \cite{bardow2016simultaneous} proposed an approach to simultaneously estimate the brightness and the the optical flow of the recorded scene: the optical flow was shown to be necessary to correctly recover sharp edges, especially in presence of fast camera movements. \\
In~\cite{reinbacher2016real} and its extended version \cite{munda2018real}, a manifold regularization method was used to reconstruct intensity images. However, predicted images exhibit significant visual artifacts and a relatively high noise. \cite{munda2018real} and the method proposed by \cite{KimHBID14} 
show best visual results under limited camera or subject movements.

Brandli \cite{brandli2014real} investigated the video decompression task and proposed an online event-based method that relies on an optimization algorithm. The synthesized image is reset with every new frame to limit the growth of the integration error. 
In \cite{scheerlinck2018continuous}, a continuous-time intensity estimation using event data is introduced. This method is based on a complementary filter, which is able to exploit both intensity frames and asynchronous events to output the gray-level image.\\
We point out that, as highlighted in \cite{scheerlinck2018continuous}, optimitazion- and filter-based methods imply the tuning of several parameters (such as the contrast threshold and the event-rate) for each recording scenario. This could limit the usability and the generalization capabilities of those methods.
In fact, in~\cite{scheerlinck2018continuous} these parameters are tuned for each testing sequence in order to improve the intensity estimation.

Recently, in \cite{pini2019video} an encoder-decoder architecture has been proposed to synthesize only gray-level frames starting from event data. The proposed approach is limited since neither an adversarial approach nor color frame information have been exploited to improve the final result.

In the method proposed by \cite{rebecq2019events}, a deep learning-based architecture is presented in order to reconstruct gray-level frames directly from event data, represented as a continuous stream of events along the acquisition time. The use of raw event data makes this method difficult to compare to the ours.

\section{\uppercase{Mathematical Formulation}}
In this section, we present definitions and mathematical notations for events and event frames, followed by their relation to intensity images and the formulation of the proposed task, \textit{i.e.}~the intensity frame synthesis. 
\begin{figure*}[t!] 
\begin{center}
  \includegraphics[width=0.98\linewidth]{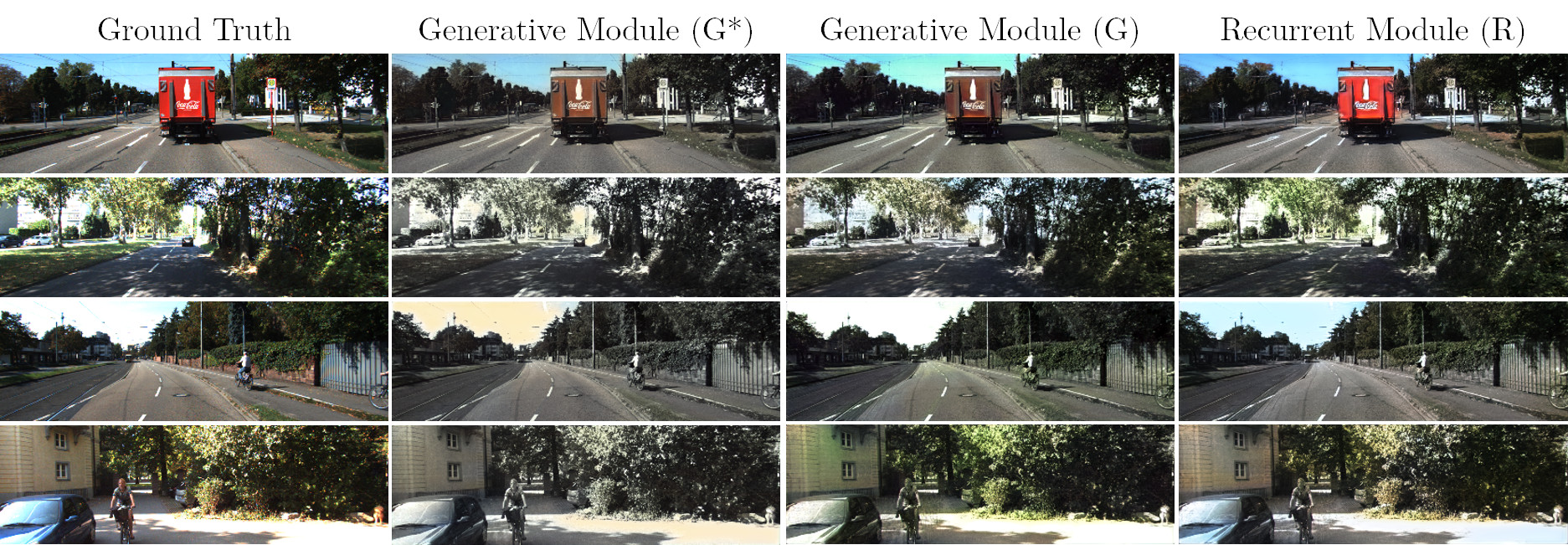}
\end{center}
  \caption{Sample output frames from \textit{Kitti} dataset~\cite{Geiger2013IJRR}. The ground truth is placed on the first column, then the output of the \textit{Generative Module} without (G*) and with (G) the discriminator (cfr. Section~\ref{subsec:generative}) and finally the output of the \textit{Recurrent Module} (R), that is able to preserve more realistic colors, enhance contrast and reduce visual artifacts. }
\label{fig:kitti}
\end{figure*}
\subsection{Event Frames}
Following the notation proposed in \cite{maqueda2018event}, the $k$-th event $e_{k}$ captured by an event camera can be represented as:
\begin{equation}
	\label{eq:event-definition}
	e_{k} = (x_{k}, y_{k}, t_{k}, p_{k})
\end{equation}
where $x_{k}$, $y_{k}$, and $t_{k}$ are the spatio-temporal coordinates of a brightness change and $p_{k} \in \{-1, +1\}$ specifies the polarity of this change, which can be either positive or negative.

By summing up all events captured in a time interval $\Delta t = [t, t+\tau]$ at a pixel-wise level, an event frame $\Phi_{\tau}(t)$ is obtained, integrating all the events occurred in that time interval.
Formally, an event frame can be defined as:
\begin{equation}
    \Phi_{\tau}(t) = \sum_{e_k \in E_{t, \, \tau}} p_k
\end{equation}

\noindent where $E_{t,\tau} = \left\{e_k \,|\, t_k \in [t, \ t+\tau] \right\}$.
Therefore, an event frame could be represented as a gray-level image of size $1 \times w  \times h$, which summarizes all events occurred in a certain time interval in a single channel. For numerical reasons, $\Phi_{\tau}(t)$ saturates if the amount of events exceeds the number of gray levels used to represent the event frame image.

\subsection{Intensity Frame Synthesis}  \label{subsec:math-framework}
The core of the proposed approach consists in learning a parametric function 
\begin{equation}
    \Gamma: \, \mathbb{R}^{c \times w \times h} \times \ \mathbb{R}^{1 \times w \times h} \, \xrightarrow{} \, \mathbb{R}^{c \times w \times h}
\end{equation}
that takes as input a $c$-channel intensity image $I_{t} \in \mathbb{R}^{c \times w\times h}$ captured at time $t$ and an event frame  $\Phi_{\tau}(t) \in \mathbb{R}^{1 \times w\times h}$, which combines pixel-level brightness changes between times $t$ and $t+\tau$, and outputs the predicted intensity image $\tilde I(t+\tau) \in \mathbb{R}^{c \times w \times h}$ at time $t+\tau$.
Here, $w$ and $h$ represent the width and the height of both intensity images and event frames.\\
It follows that
\begin{equation}
	\label{eq:framesynthesis}
	{\tilde I(t+\tau)} = \Gamma(I(t), \ \Phi_{\tau}(t), \ \theta)
\end{equation}
where $\theta$ corresponds to the parameters of the function $\Gamma$,
that we define as the combination of multiple parametric functions (Sec.~\ref{subsec:network-architecture}).

\subsection{Difference of Images as Event Frames} \label{sec:diffasevent}
Event cameras are naturally triggered by pixel-level logarithmic brightness changes and thus they provide some output data only if there is a relative movement between the sensor and the objects in the scene or a brightness change occurs~\cite{gehrig2018asynchronous}.
For small time intervals, \textit{i.e.}~small values of $\tau$, the brightness variation can be approximated with a first-order Taylor's approximation as:
\begin{equation}
    \lim_{\tau \rightarrow 0} \frac{\delta L}{\delta t} \tau \approx L(t+\tau) - L(t) \doteq \Delta L
\end{equation}
where $L(t) = \log\left(Br\left(I(t)\right)\right)$, $I(t)$ is the image acquired at time $t$ and $Br(\cdot)$ is a function to convert a $c$-channel image into the corresponding single-channel brightness. In the experiments, for instance, RGB images are converted into brightness images using the standard channel weights defined as $[0.299, 0.587, 0.114]$.
\\
Therefore, an event frame $\Phi_{\tau}(t)$ can be approximated as follows:
\begin{equation}
	\label{eq:eventasdiff}
	\Phi_{\tau}(t) \approx \Delta L = 
	\log\left[ Br\left(I (t+\tau) \right) \right] - 
	\log\left[ Br\left(I (t) \right) \right]
\end{equation}
Thanks to this assumption, given two intensity frames $I(t)$ and $I(t+\tau)$, it is possible to retrieve the corresponding event frame $\Phi_{\tau}(t)$ for small values of $\tau$. However, since intensity frames have more than one channel (\eg, three channels for RGB images), $I(t+\tau)$ cannot be analytically obtained given $I(t)$ and $\Phi_{\tau}(t)$.

\section{\uppercase{Implementation}} \label{subsec:network-architecture}
An overview of the proposed architecture is depicted in Figure \ref{fig:overview}. 
The framework integrates two main components.
The first one -- the \textit{Generative Module} (G) -- receives an intensity image $I(t)$ and an event frame $\Phi_{\tau}(t)$ as input and synthesizes the frame $I(t+\tau)$ as output.\\
The second one -- the \textit{Recurrent Module} (R) -- refines the output of the Generative component, relying on the temporal coherence of a sequence of frames.

\subsection{Generative Module} \label{subsec:generative}
We follow the conditional GAN paradigm~\cite{mirza2014conditional,isola2017image,borghi2018face} for the designing of the \textit{Generative Module}. \\
The module consists of a generative network $G$ and a discriminative network $D$~\cite{goodfellow2014generative,mirza2014conditional}. 
Exploiting the \textit{U-Net} architecture~\cite{ronneberger2015u}, $G$ is defined as a fully-convolutional deep neural network with skip connections between layers $i$ and $n-i$, where $n$ is the total number of layers. The discriminative network proposed by~\cite{isola2017image} is employed as $D$.\\
In formal terms, $G$ corresponds to an estimation function that predicts the intensity frame $\tilde I(t+\tau) = G\left(I(t) \oplus \Phi_{\tau}(t)\right)$ from the concatenation of an intensity frame and an event frame at time $t$  (cfr. Equation~\ref{eq:framesynthesis}) while $D$ corresponds to a discriminative function able to distinguish between real and generated frames.\\
The training procedure can be formalized as the optimization of the following min-max problem:
\begin{align}
	\label{eq:minmax}
\begin{aligned}
 	\min_{\theta_{G}} \max_{\theta_{D}}\ &\mathbb{E}_{x \sim p(x), y \sim p(y)}[\log D(x, y)] \\
 	&+ \mathbb{E}_{x \sim p(x)}[\log (1 - D(x, G(x)))]
\end{aligned}
\end{align}
\noindent where $D(x, y)$ is the probability of being a \textit{real} frame and $1 - D(x,G(x))$ is the probability to be a synthesized frame, $p(x)$ is the distribution of concatenated frames $I(t) \oplus \Phi_{\tau}(t)$, and $p(y)$ is the distribution of frames $\tilde I(t+\tau)$.\\
This approach leads to a \textit{Generative Module} which is capable of translating pixel intensities accordingly to an event frame and producing output frames that are visually similar to the real ones.

\begin{table*}[th!]
\centering
\small
\caption{Comparison between the proposed method and state of the art approaches using per-pixel and perceptual metrics.}
\setlength\tabcolsep{2pt}
\resizebox{1\textwidth}{!}{
\begin{tabular}{cc c C{9mm}C{9mm} c C{9mm}C{9mm}C{9mm} c C{9mm}C{9mm}C{9mm} c C{9mm}C{9mm} c C{18mm}}
\toprule
\multirow{2}{*}{\textbf{Dataset}} & \multirow{2}{*}{\textbf{Method}} & &\multicolumn{2}{c}{\textbf{Norm} $\downarrow$}& &\multicolumn{3}{c}{\textbf{RMSE} $\downarrow$} &  &\multicolumn{3}{c}{\textbf{Threshold} $\uparrow$} & &\multicolumn{2}{c}{\textbf{Indexes} $\uparrow$} & &\textbf{Perceptual} $\downarrow$ \\
&  & &$L_1$ &$L_2$& &Lin &Log &Scl & &$1.25$ &$1.25^2$ &$1.25^3$  & &PSNR  &SSIM & &LPIPS  \\
\midrule
\multirow{4}{*}{DDD17}
                        & Munda et al.  & & 0.268 & 94.277 & & 0.314 & 5.674 & 5.142 & & 0.152 & 0.448 & 0.536 & & 10.244 & 0.216 & & 0.637  \\
                        & Scheerlinck et al.   & & 0.080 & 29.249 & & 0.098 & 4.830 & 4.352 & & 0.671 & 0.781 & 0.827 & & 20.542 & 0.702 & & 0.208  \\
                        & Pini et al.   & & 0.027 & 8.916 & & 0.040 & 4.048 & 3.571 & & 0.775 & 0.848 & 0.875 & & 29.176 & 0.864 & & \textbf{0.105}  \\
                        & \textbf{Ours}   & & \textbf{0.022} & \textbf{8.583} & & \textbf{0.039} & \textbf{3.766} & \textbf{3.408} & & \textbf{0.787} & \textbf{0.855} & \textbf{0.880} & & \textbf{29.428} & \textbf{0.884} & & 0.107 \\
                    
                    \midrule

\multirow{4}{*}{MVSEC}
                        & Munda et al.  & & 0.160 & 86.419 & & 0.288 & 8.985 & 8.016 & & 0.088 & 0.163 & 0.232 & & 11.034 & 0.181 & & 0.599  \\
                        & Scheerlinck et al.   & & 0.067 & 26.794 & & 0.089 & 7.313 & 6.982 & & 0.263 & 0.357 & 0.467 & & 21.070 & 0.551 & & 0.257 \\
                        & Pini et al.   & & 0.026 & 12.062 & & 0.054 & \textbf{6.443} & 6.102 & & \textbf{0.525} & \textbf{0.642} & \textbf{0.708} & & 25.866 & 0.740 & & 0.172 \\
                        & \textbf{Ours}   & & \textbf{0.022} & \textbf{11.216} & & \textbf{0.051} & 6.559 & \textbf{6.003} & & 0.514 & 0.637 & 0.699 & & \textbf{26.366} & \textbf{0.845} & & \textbf{0.137} \\

                        \bottomrule
\end{tabular}
}
\label{tab:competitors}
\end{table*}

\subsection{Recurrent Module}
The architecture of the \textit{Recurrent Module} is a combination of an \textit{encoder-decoder} architecture and a \textit{Convolutional LSTM} (ConvLSTM) module~\cite{xingjian2015convolutional}.
The underlying idea is that while the \textit{Generative Module} learns how to successfully combine intensity and event frames, the \textit{Recurrent Module}, capturing the context of the scene and its temporal evolution, learns to visually refine the synthesized frames, removing artifacts, enhancing colors, and improving the temporal coherence.\\
We adopt the same \textit{U-Net} architecture of the \textit{Generative Module} and we insert a $512$-channel two-layer ConvLSTM block in the middle of the hourglass model.
During the training phase, the \textit{Recurrent Module} receives as input a sequence of frames produced by the \textit{Generative Module} and outputs a sequence of the same length, sequentially updating the internal state.
The activation of each ConvLSTM layer can be defined as follows:
\begin{align}
I_s &= \sigma(W_i \ast X_s + U_i \ast H_{s-1} + b_i) \\
F_s &= \sigma(W_f \ast X_s + U_f \ast H_{s-1} + b_f) \\
O_s &= \sigma(W_o \ast X_s + U_o \ast H_{s-1} + b_o) \\
G_s &= tanh(W_c \ast X_s + U_c \ast H_{s-1} + b_c)   \\
C_s &= F_s \odot C_{s-1} + I_s \odot G_s   \\
H_s &= O_s \odot tanh(C_s)
\end{align}
where, $I_s$, $F_s$, $O_s$ are the gates, $C_s, C_{s-1}$ are the memory cells, $G_s$ is the candidate memory, and $H_s$, $H_{s-1}$ are the hidden states. 
Each $b$ is a learned bias, each $W$ and $U$ are a learned convolutional kernel, and $X_s$ corresponds to the input.
Finally, $\ast$ represents the convolutional operator while $\odot$ is the element-wise product. %

\subsection{Training Procedure} \label{subsec:training}
The framework is trained in two consecutive steps.

In the first phase, the \textit{Generative Module} $G$ is trained following the adversarial approach detailed in Section \ref{subsec:generative}.
We optimize the network using Adam~\cite{KingmaB14} with learning rate $0.0002$, $\beta_1 = 0.5$, $\beta_2 = 0.999$, and a batch size of $8$. In order to improve the stability of the training process, the discriminator is updated every $8$ training steps of the generator.
The objective function of $D$ is the common binary categorical cross entropy loss, while the objective function of $G$ is a weighted combination of the adversarial loss (\textit{i.e.}~the binary crossentropy) and the \textit{Mean Squared Error} (MSE) loss.

In the second phase, the \textit{Recurrent Module} $R$ is trained while keeping the parameters of the \textit{Generative Module} fixed.
We apply the Adam optimizer with the same hyper-parameters we used for the \textit{Generative Module}, with the exception of the batch size which is set to $4$.
The objective function of the module is a weighted combination of the MSE loss and the \textit{Structural Similarity} index (SSIM) loss which is defined as:
\begin{equation}
	\label{eq:rec_loss}
	\text{SSIM}(p,q) = \frac{(2\mu_p\mu_q + c_1)(2\sigma_{pq} + c_2)}{(\mu_p^2 + \mu_q^2 + c_1)(\sigma_p^2 + \sigma_q^2 + c_2)}
\end{equation}
Given two windows $p$, $q$ of equal size, $\mu_{p,q}$, $\sigma_{p,q}$ are the mean and variance of $p,q$ while $c_{1,2}$ are used to stabilize the division. 
See~\cite{wang2004image} for further details.
The losses are combined with a weight of $0.5$ each.
The network is trained with a fixed sequence length, which corresponds to the length of the sequences used during the evaluation phase.\\
Only during the testing phase, to obtain a sequence of synthesized frames, the framework receives as input the previously-generated images or an intensity key-frame.

\section{\uppercase{Framework Evaluation}}
In this section, we present the datasets that we used to train and test the proposed framework.
Then, we describe the evaluation procedure that has been employed to assess the quality of the synthesized frames, followed by the report of the experimental results and their analysis.

\begin{table*}[t]
\centering
\small
\caption{Experimental results of pixel-wise metrics computed on synthesized frames from \textit{DDD17}, \textit{MVSEC}, \textit{Kitti}, and \textit{Cityscapes} (CS) datasets.
Details on adopted metrics are reported in Section \ref{sec:metrics}.
Tests are carried out employing the \textit{Generative Module} (G), and both the \textit{Generative} and  \textit{Recurrent Module} (G+R).}
\setlength\tabcolsep{3pt}
\resizebox{1\textwidth}{!}{
\begin{tabular}{cc c C{9mm}C{9mm} c C{9mm}C{9mm} c C{9mm}C{9mm}C{9mm} c C{9mm}C{9mm}C{9mm} c C{9mm}C{9mm}}
\toprule
\multirow{2}{*}{\textbf{Dataset}} & \multirow{2}{*}{\textbf{Model}} & &\multicolumn{2}{c}{\textbf{Norm} $\downarrow$} & &\multicolumn{2}{c}{\textbf{Difference} $\downarrow$}  & &\multicolumn{3}{c}{\textbf{RMSE} $\downarrow$} &  &\multicolumn{3}{c}{\textbf{Threshold} $\uparrow$} & &\multicolumn{2}{c}{\textbf{Indexes}$\uparrow$}  \\
&  & &$L_1$ &$L_2$ & &Abs  &Sqr  & &Lin &Log &Scl & &$1.25$ &$1.25^2$ &$1.25^3$  & &PSNR  &SSIM\\
\midrule
\multirow{2}{*}{DDD17}  
                        &G       & & 0.029 & 9.658 & & 0.114 & 0.007 & & 0.044 & 2.296 & 2.268 & & 0.854 & 0.919 & 0.941 & & 28.486 & 0.876 \\
                        &G+R     & & 0.022 & 8.583 & & 0.167 & 0.006 & & 0.039 & 3.766 & 3.408 & & 0.787 & 0.855 & 0.880 & & 29.428 & 0.884 \\

                        \midrule

\multirow{2}{*}{MVSEC}  
                        &G       & & 0.026 & 12.830 & & 0.311 & 0.013 & & 0.058 & 6.302 & 6.233 & & 0.562 & 0.675 & 0.733 & & 25.309 & 0.784 \\
                        &G+R     & & 0.022 & 11.216 & & 0.354 & 0.010 & & 0.051 & 6.559 & 6.003 & & 0.514 & 0.637 & 0.699 & & 26.366 & 0.845 \\

                        \midrule

\multirow{2}{*}{Kitti}  
                        &G    & &0.030           &10.95              & &0.125           &0.006               & &0.048           &0.472           &0.463           & &0.782           &0.940            &0.981           & &27.140          &0.919\\
                        &G+R  & &0.029  &10.71     & &0.105  &0.005      & &0.046  &0.194  &0.191  & &0.846  &0.968   &0.991  & &27.295 &0.928\\

                        \midrule

\multirow{2}{*}{CS}     
                        &G    & &0.019           &4.534              & &0.086           &0.003               & &0.025           &0.232           &0.211           & &0.877           &0.974            &0.992           & &32.769          &0.962\\
                        &G+R  & &0.015  &4.192     & &0.059  &0.002      & &0.023  &0.172  &0.170  & &0.968  &0.997   &0.999  & &33.315 &0.971\\

                        \bottomrule
\end{tabular}
}
\label{tab:pixel-wise-metrics}
\end{table*}

\subsection{Datasets}
Due to the recent commercial release of event cameras, only few event-based datasets are currently publicly-released and available in the literature.
These datasets still lack the data variety and the annotation quality which is common for RGB datasets.
These considerations have motivated us to exploit the mathematical intuitions presented in Section~\ref{sec:diffasevent} in order to take advantage of non-event public automotive datasets~\cite{Geiger2013IJRR,cordts2016cityscapes}, which are richer in terms of annotations and data quality, along with two recent event-based automotive datasets~\cite{binas2017ddd17,zhu2018multivehicle}.

\noindent \textbf{DDD17.}
Binas \etal~\cite{binas2017ddd17} introduced \textit{DDD17: End-to-end DAVIS Driving Dataset}, which is the first open dataset of annotated event driving recordings. 
The dataset is captured by a \textit{DAVIS} sensor~\cite{brandli2014240} and includes both gray-level frames ($346 \times 260$ px) and event data.
Sequences are captured in urban and highway scenarios, during day and night and under different weather conditions. 
Similar to~\cite{maqueda2018event}, experiments are carried out selecting only 
sequences labelled as \texttt{day}, \texttt{day} \texttt{wet}, and \texttt{day} \texttt{sunny}, but we create train, validation, and test split using different sequences.

\noindent \textbf{MVSEC.}
The \textit{Multi Vehicle Stereo Event Camera Dataset}~\cite{zhu2018multivehicle} contains data acquired from four different vehicles, in both indoor and outdoor environments, during day and night, using a pair of \textit{DAVIS 346B} event cameras ($346 \times 260$ px), a stereo camera, and a \textit{Velodyne} lidar.
In this paper, we use only the outdoor car scenes recorded during the day. From these, we select the first $70\%$ as train set, and the following as validation ($10\%$) and test ($20\%$) set.

\noindent \textbf{Kitti.}
The \textit{Kitti Vision Benchmark Suite} was introduced in~\cite{Geiger2012CVPR}.
In this work, we use the \textit{KITTI raw}~\cite{Geiger2013IJRR} subset, which includes 6 hours of $1242 \times 375$ rectified RGB image sequences captured on different road scenarios with a temporal resolution of $10$Hz. 
The dataset is rich of annotations, as depth maps and semantic segmentation.
We adopt the train and the validation split proposed in~\cite{Uhrig2017THREEDV} to respectively train the method and to validate and test it.

\noindent \textbf{Cityscapes.}
The \textit{Cityscapes} dataset~\cite{cordts2016cityscapes} consists of thousands of RGB frames with a high spatial resolution ($2048 \times 1024$ px) and shows varying and complex scene layouts and backgrounds. 
Fine and coarse annotations of $30$ different object classes are provided as both semantic and instance-wise segmentation.
We select a particular subset, namely \texttt{leftImg8bit\_sequence}, following official splits, in order to use sequences with a frame rate of $17$Hz and to have access to fine semantic segmentation annotations.

\subsection{Metrics}\label{sec:metrics}
Inspired by~\cite{eigen2014depth,isola2017image}, we exploited a variety of metrics to check the quality of the generated images, being aware that evaluating synthesized images is, in general, a still open problem~\cite{salimans2016improved}.
We firstly design a set of experiments in order to investigate the contribution of each single module of the proposed framework and to compare it with state-of-art methods by using pixel-wise and perceptual metrics. 
Then, we exploit off-the-shelf networks pre-trained on public datasets in order to evaluate semantic segmentation and object detection scores on generated images.

\begin{figure*}[th!]
    \centering
    \begin{subfigure}[t]{0.24\textwidth}
        \centering
        \caption*{{\small DDD17}}
        \includegraphics[width=0.95\linewidth]{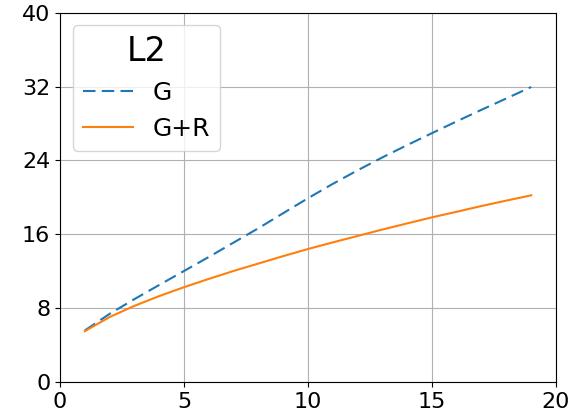}
        \par\medskip
        \includegraphics[width=0.95\linewidth]{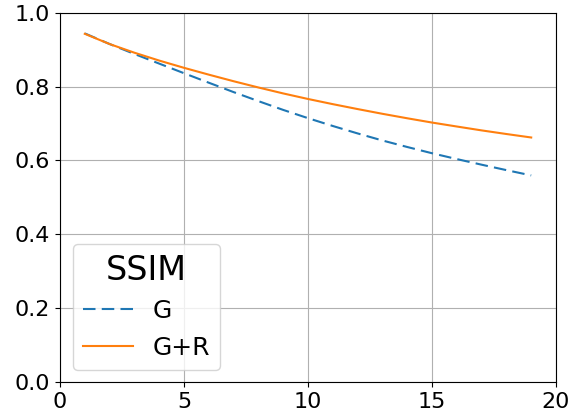}
    \end{subfigure}
    \centering
    \begin{subfigure}[t]{0.24\textwidth}
        \centering
        \caption*{{\small MVSEC}}
        \includegraphics[width=0.95\linewidth]{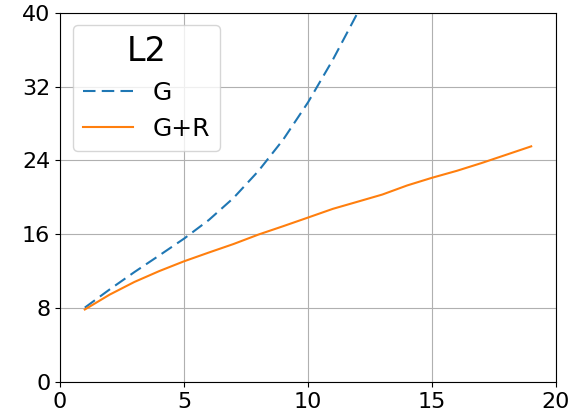}
        \par\medskip
        \includegraphics[width=0.95\linewidth]{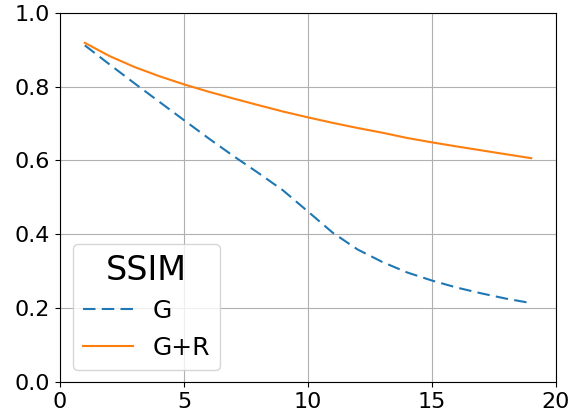}
    \end{subfigure}
    \centering
    \begin{subfigure}[t]{0.24\textwidth}
        \centering
        \caption*{{\small Kitti}}
        \includegraphics[width=0.95\linewidth]{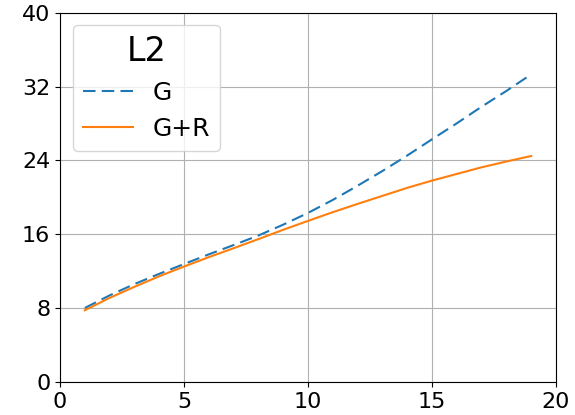}
        \par\medskip
        \includegraphics[width=0.95\linewidth]{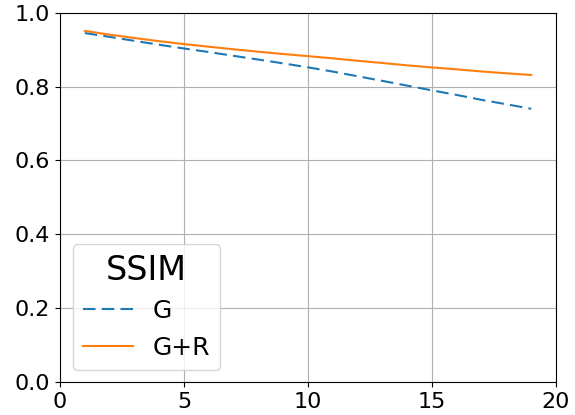}
    \end{subfigure}
    \centering
    \begin{subfigure}[t]{0.24\textwidth}
        \centering
        \caption*{{\small Cityscapes}}
        \includegraphics[width=0.95\linewidth]{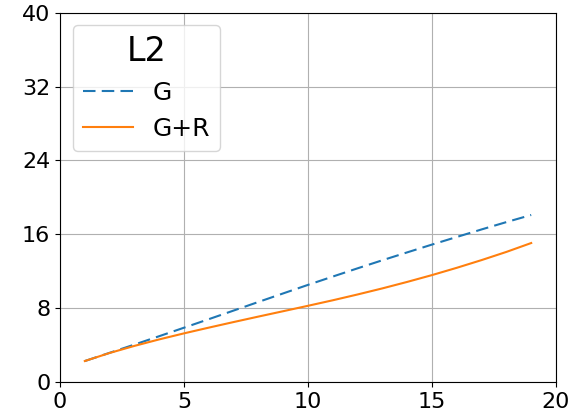}
        \par\medskip
        \includegraphics[width=0.95\linewidth]{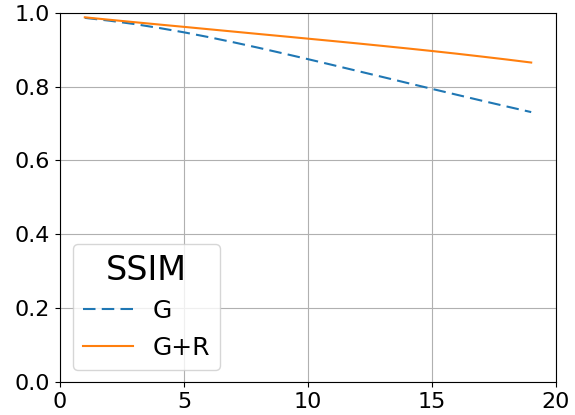}
    \end{subfigure}
    \caption{Variation of $L_2$ and \textit{SSIM} as a function of the $i$-th synthesized frame since the last key-frame by the \textit{Generative Module}, without (blue) and with (orange) the discriminator, and the \textit{Recurrent Module} (green), computed on \textit{DDD17}, \textit{MVSEC}, \textit{Kitti}, and \textit{Cityscapes}. The horizontal axis refers to the frame on which the metric is calculated, starting from an initial color frame and estimating the following ones.}
    \label{fig:graphs_metrics}
\end{figure*}

\noindent \textbf{Pixel-wise and perceptual metrics.}
A collection of evaluation metrics is used to assess the quality of the synthesized images.
In particular, we report the $L1$ and $L2$ distance, the root mean squared error (RMSE), and the percentage of pixel under a certain error threshold ($\delta$-metrics).
Moreover, we include the \textit{Peak Signal-to-Noise Ratio} (PSNR), which estimates the level of noise in logarithmic scale, and the \textit{Structural Similarity} (SSIM)~\cite{wang2004image}, which measures the perceived closeness between two images.
Finally, the visual quality of the generated images is assessed through the \textit{Learned Perceptual Image Patch Similarity} (LPIPS)~\cite{zhang2018perceptual} which was shown to correlate well with human judgement~\cite{zhang2018perceptual}.

\noindent \textbf{Semantic segmentation score.}
We adopt a pre-trained semantic classifier to measure the accuracy of a certain set of pixels to be a particular class. Specifically, we rely on the validation set of the \textit{Kitti} and \textit{Cityscapes} dataset.
If synthesized images are close to the real ones, the classifier will achieve a comparable accuracy to the one obtained on the reference dataset.
We adopt the recent state-of-art \texttt{WideResNet+38+DeepLab3}~\cite{rotabulo2017place} trained on the original train annotations of the \textit{Cityscapes} dataset.
Since semantic fine annotations are provided only for a limited subset of frames in each sequence, we compare these annotations with the semantic maps produced using as input the last frame of a synthesized sequence (\textit{i.e.}~the worst case).

\noindent \textbf{Object detection score.}
A pre-trained object detector is employed in order to investigate if the proposed model is able to preserve details, locations, and realistic aspect of the objects that appear in the scene.
We adopt the popular \texttt{Yolo-v3} network~\cite{yolov3}, a real-time state-of-the-art object detection system, pre-trained on the \textit{COCO} dataset~\cite{lin2014microsoftcoco}.
In this way, since we use automotive datasets, we investigate the ability of the proposed framework to preserve objects in the generated frames, in particular people, trucks, cars, buses, trains, and stop signals.

\subsection{Experimental Results}
 For a fair comparison, we empirically set the same sequence length of $6$ synthesized frames for every experiment and competitor method reported in this section.
We split data following the training and testing sets of each dataset, and we use the validation set to stop the training procedure.
When considering \textit{DDD17} and \textit{MVSEC}, only real event data are used while we obtain synthetic event frames on \textit{Kitti} and \textit{Cityscapes}.
We adapt the image resolution of the original data to comply with the \textit{U-Net} architecture (see Sec.~\ref{subsec:generative}) requirements while trying to keep the original image aspect ratio. 
Therefore, we adopt input images with a spatial resolution of $416 \times 128$ for \textit{Kitti}, $256 \times 128$ for \textit{Cityscapes}, and $256 \times 192$ for \textit{DDD17} and \textit{MVSEC}.

\begin{figure*}[th!]
\captionsetup[subfigure]{labelformat=empty}
\centering
\resizebox{0.98\textwidth}{!}{
    \centering
    \begin{subfigure}[t]{0.19\textwidth}
        \centering
        \caption{Ground Truth}
        \includegraphics[width=0.95\linewidth]{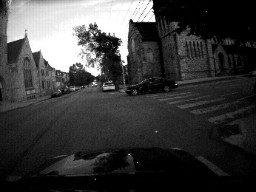}\vspace{1px}
        \vspace{3px}
        \includegraphics[width=0.95\linewidth]{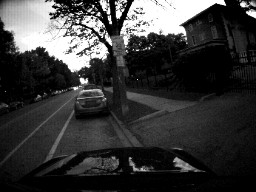}
        \vspace{1px}
        \includegraphics[width=0.95\linewidth]{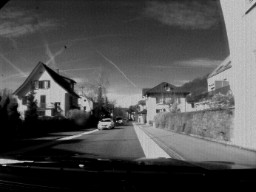}
        \includegraphics[width=0.95\linewidth]{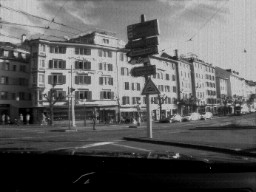}
    \end{subfigure}
    \centering
    \begin{subfigure}[t]{0.19\textwidth}
        \centering
        \caption{Event Frame}
        \includegraphics[width=0.95\linewidth]{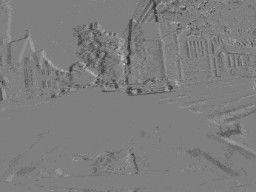}\vspace{1px}
        \vspace{3px}
        \includegraphics[width=0.95\linewidth]{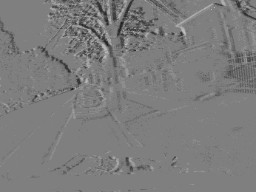}
        \vspace{1px}
        \includegraphics[width=0.95\linewidth]{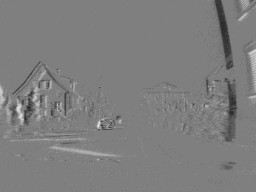}
        \includegraphics[width=0.95\linewidth]{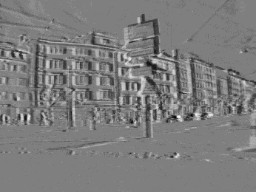}
    \end{subfigure}
    \centering
    \begin{subfigure}[t]{0.19\textwidth}
        \centering
        \caption{Munda et al.}
        \includegraphics[width=0.95\linewidth]{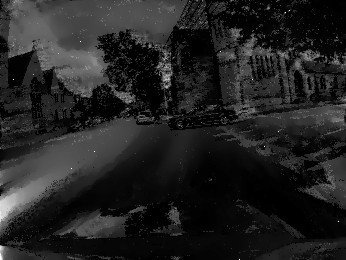}\vspace{1px}
        \vspace{3px}
        \includegraphics[width=0.95\linewidth]{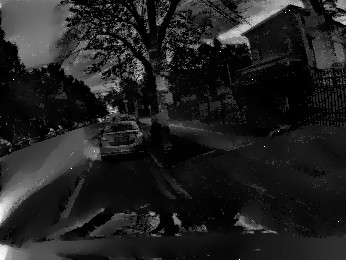}
        \vspace{1px}
        \includegraphics[width=0.95\linewidth]{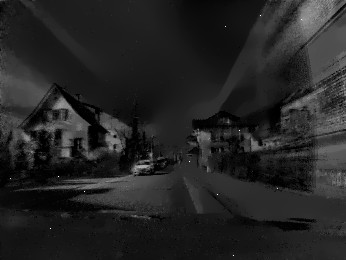}
        \includegraphics[width=0.95\linewidth]{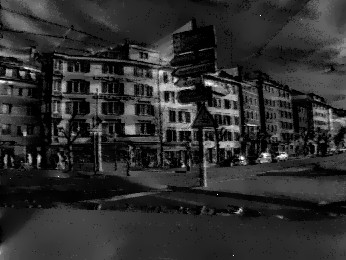}
    \end{subfigure}
    \begin{subfigure}[t]{0.19\textwidth}
        \centering
        \caption{Scheerlinck et al.}
        \includegraphics[width=0.95\linewidth]{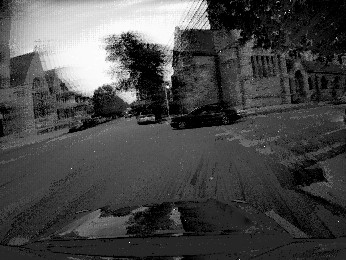}\vspace{1px}
        \vspace{3px}
        \includegraphics[width=0.95\linewidth]{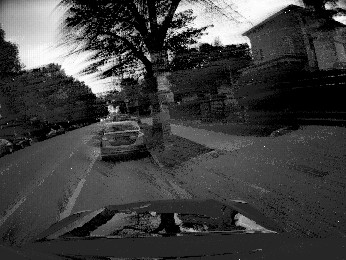}
        \vspace{1px}
        \includegraphics[width=0.95\linewidth]{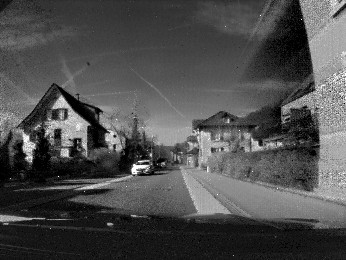}
        \includegraphics[width=0.95\linewidth]{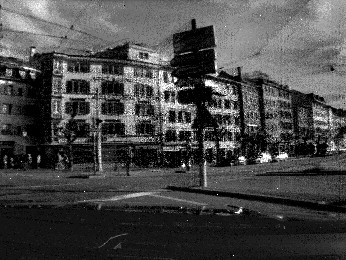}
    \end{subfigure}
    \begin{subfigure}[t]{0.19\textwidth}
        \centering
        \caption{\textbf{Ours}}
        \includegraphics[width=0.95\linewidth]{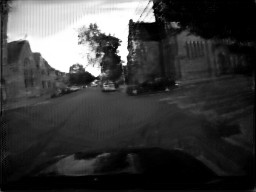}\vspace{1px}
        \vspace{3px}
        \includegraphics[width=0.95\linewidth]{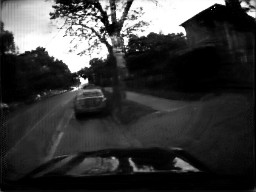}
        \vspace{1px}
        \includegraphics[width=0.95\linewidth]{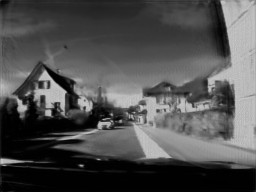}
        \includegraphics[width=0.95\linewidth]{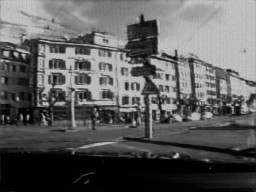}
    \end{subfigure}
}
    \caption{Output of the proposed algorithm compared to existing methods on MVSEC and DDD17 datasets.
    }
    \label{fig:final}
\end{figure*}

In Table~\ref{tab:competitors}, per-pixel evaluation shows that our model overcomes all the competitors on both the event-based \textit{DDD17} and \textit{MVSEC} datasets.
We point out that \cite{munda2018real} is based only on event data while the input of \cite{scheerlinck2018continuous,pini2019video} are both gray-level images and events, similarly to our method.
According to Table~\ref{tab:competitors}, the visual results reported in Figure~\ref{fig:final} confirm the superior quality of the images synthesized by our method and suggest that the proposed learning-based approach can be an alternative of filter-based algorithms.
Indeed, visual artifacts (\eg shadows) and high noise (\eg salt and pepper) are visible in the competitor generated frames while our method produces more accurate brightness levels. In addition, competitors are limited to the gray-level domain only.

\begin{table}[t!]
\centering
\small
\caption{Semantic Segmentation and Object Detection scores computed on synthesized frames from \textit{Kitti} and \textit{Cityscapes}. 
Results are compared with the Ground Truth (GT), when available. }
\setlength\tabcolsep{4pt}
\resizebox{1\columnwidth}{!}{
\begin{tabular}{cc C{0mm} ccc C{1mm} cc}
\toprule
\multirow{2}{*}{\textbf{Data}} & \multirow{2}{*}{\textbf{Model}}  & &\multicolumn{3}{c}{\textbf{Semantic Segmentation} $\uparrow$} & &\multicolumn{2}{c}{\textbf{Object Det.} $\uparrow$}  \\
%
&    & &\footnotesize{Per-pixel} &\footnotesize{Per-class} &\footnotesize{class IoU}    & & \footnotesize{mIoU} &\footnotesize{\%}    \\ 
\midrule
\multirow{3}{*}{Kitti}    &G      & &0.814 &0.261 &0.215    & &0.914   &65.8 \\ 
                                                &G+R    & &0.813 &0.261 &0.215    & &0.912   &71.4 \\ 
                                                &GT 	& &0.827 &0.283 &0.235    & &-       &- \\ 

        \midrule 

\multirow{3}{*}{CS}         &G      & &0.771      &0.197     &0.162     & &0.924   &83.5       \\ 
                                                        &G+R 	& &0.790      &0.201     &0.166     & &0.926   &86.4       \\ 
                                                        &GT 	& &0.828      &0.227     &0.192     & &-      &-   \\ 
                                
                                \bottomrule
\end{tabular}
}
\label{tab:ssod-metrics}
\end{table}

As an ablation study, we exploit the pixel-wise metrics also to understand the contribution of each single module of the proposed system. Table~\ref{tab:pixel-wise-metrics} and Figure~\ref{fig:kitti} show
that the output of the \textit{Generative Module} has a good level of quality and learns efficiently to alter pixel values accordingly to event frames. \textit{Recurrent Module} visually improves the output frames, 
enhancing the colors, the level of details, and the temporal coherence. \\
Generally, we note that the low quality of the gray-level images provided in \textit{DDD17} and \textit{MVSEC} partially influences the performance of the framework.

\begin{figure*}[h!]
\captionsetup[subfigure]{labelformat=empty}
\centering
\resizebox{1\textwidth}{!}{
    \centering
    \begin{subfigure}[t]{0.19\textwidth}
        \centering
        \caption{{\small Original}}
        \includegraphics[width=0.98\linewidth]{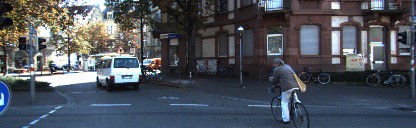}
        \includegraphics[width=0.98\linewidth]{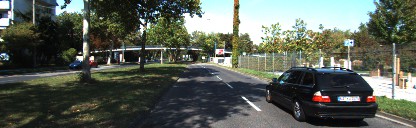}
        \includegraphics[width=0.98\linewidth]{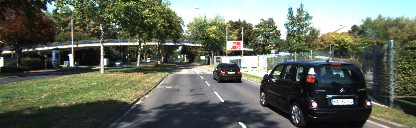}
        \includegraphics[width=0.98\linewidth]{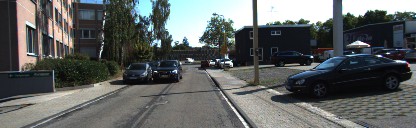}
        \includegraphics[width=0.98\linewidth]{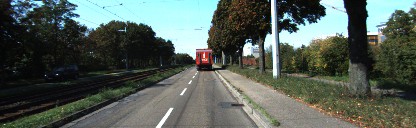}
        \includegraphics[width=0.98\linewidth]{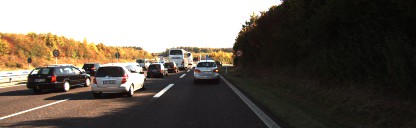}
    \end{subfigure}
    \centering
    \begin{subfigure}[t]{0.19\textwidth}
        \centering
        \caption{{\small Ground Truth}}
        \includegraphics[width=0.98\linewidth]{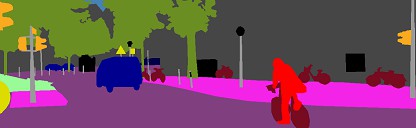}
        \includegraphics[width=0.98\linewidth]{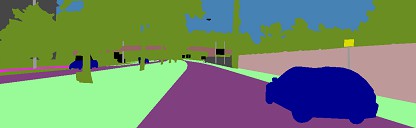}
        \includegraphics[width=0.98\linewidth]{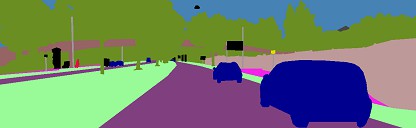}
        \includegraphics[width=0.98\linewidth]{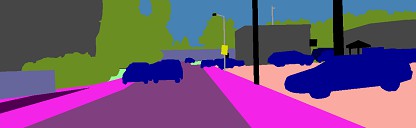}
        \includegraphics[width=0.98\linewidth]{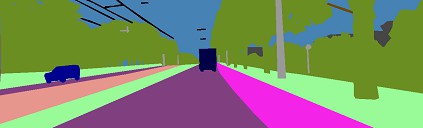}
        \includegraphics[width=0.98\linewidth]{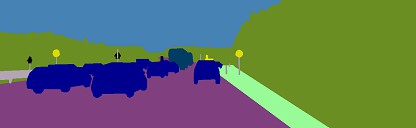}
    \end{subfigure}
    \centering
    \begin{subfigure}[t]{0.19\textwidth}
        \centering
        \caption{{\small SS on Original}}
        \includegraphics[width=0.98\linewidth]{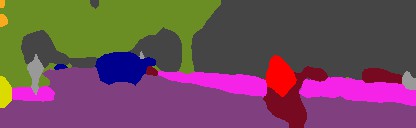}
        \includegraphics[width=0.98\linewidth]{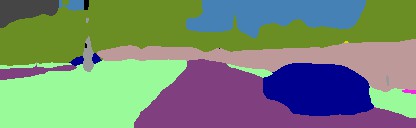}
        \includegraphics[width=0.98\linewidth]{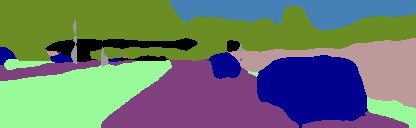}
        \includegraphics[width=0.98\linewidth]{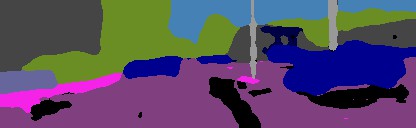}
        \includegraphics[width=0.98\linewidth]{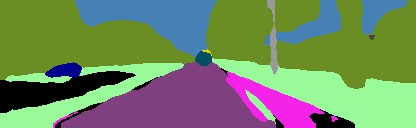}
        \includegraphics[width=0.98\linewidth]{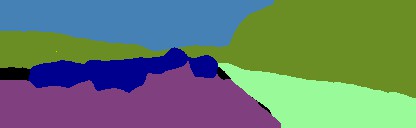}
    \end{subfigure}
    \begin{subfigure}[t]{0.19\textwidth}
        \centering
        \caption{{\small SS on Ours}}
        \includegraphics[width=0.98\linewidth]{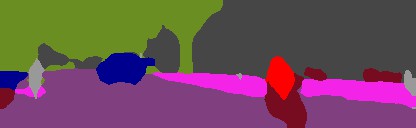}
        \includegraphics[width=0.98\linewidth]{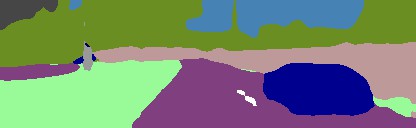}
        \includegraphics[width=0.98\linewidth]{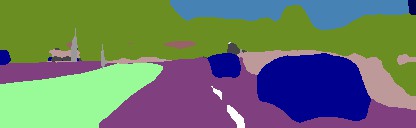}
        \includegraphics[width=0.98\linewidth]{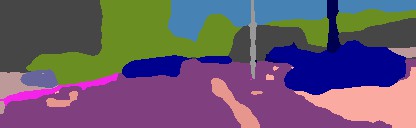}
        \includegraphics[width=0.98\linewidth]{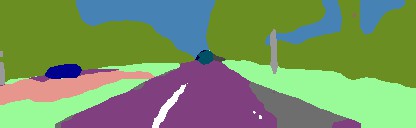}
        \includegraphics[width=0.98\linewidth]{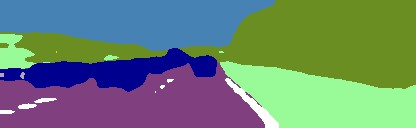}
    \end{subfigure}
}
    \caption{Semantic Segmentation (SS) applied on \textit{Kitti} dataset~\cite{Geiger2013IJRR}.
    From left, we report the original RGB frame, the ground truth of the semantic segmentation and then the segmentation computed on original and synthesised frames.}
    \label{fig:ss_kitti}
\end{figure*}

\begin{figure*}[th]
\captionsetup[subfigure]{labelformat=empty}
\centering
\resizebox{0.8\textwidth}{!}{
    \centering
    \begin{subfigure}[t]{0.19\textwidth}
        \centering
        \caption{{\small Original}}
        \includegraphics[width=0.98\linewidth]{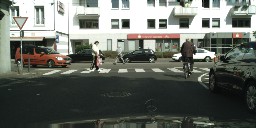}
        \includegraphics[width=0.98\linewidth]{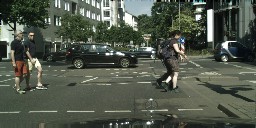}
        \includegraphics[width=0.98\linewidth]{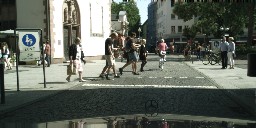}
        \includegraphics[width=0.98\linewidth]{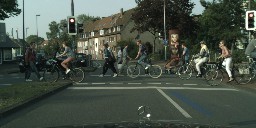}
        \includegraphics[width=0.98\linewidth]{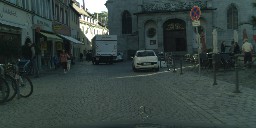}
    \end{subfigure}
    \centering
    \begin{subfigure}[t]{0.19\textwidth}
        \centering
        \caption{{\small OD on original}}
        \includegraphics[width=0.98\linewidth]{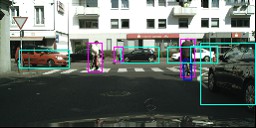}
        \includegraphics[width=0.98\linewidth]{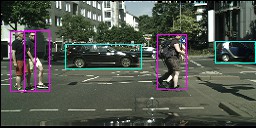}
        \includegraphics[width=0.98\linewidth]{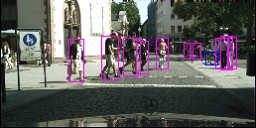}
        \includegraphics[width=0.98\linewidth]{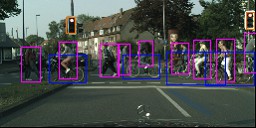}
        \includegraphics[width=0.98\linewidth]{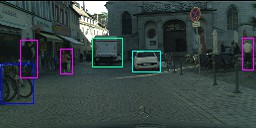}
    \end{subfigure}
    \centering
    \begin{subfigure}[t]{0.19\textwidth}
        \centering
        \caption{{\small OD on Ours}}
        \includegraphics[width=0.98\linewidth]{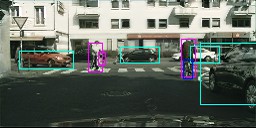}
        \includegraphics[width=0.98\linewidth]{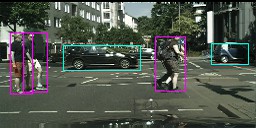}
        \includegraphics[width=0.98\linewidth]{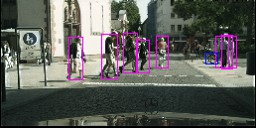}
        \includegraphics[width=0.98\linewidth]{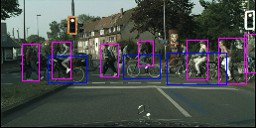}
        \includegraphics[width=0.98\linewidth]{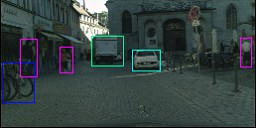}
    \end{subfigure}
}
    \caption{Object Detection (OD) on \textit{Cityscapes} dataset~\cite{cordts2016cityscapes}.}
    \label{fig:od_cs}
\end{figure*}

In Table \ref{tab:ssod-metrics}, results are reported in terms of per-pixel, per-class, and \textit{IoU} accuracy for the Semantic Segmentation score and in terms of mean \textit{IoU} and percentage of the correctly detected objects for the Object Detection score. \\
Segmentation results confirm that our approach can be a valid option to avoid the development of completely new vision algorithms relying on event data. Also in this case, the \textit{Recurrent Module} improves the final score (Fig. \ref{fig:ss_kitti}).
The Object Detection scores are reported in Table \ref{tab:ssod-metrics} in terms of mean \textit{IoU} on detection bounding boxes and the percentage of objects detected with respect to the \texttt{Yolo-v3} network detections on the ground truth images. 
Object detection scores are interesting, since we note that even though the mean \textit{IoU} computed is similar, the \textit{Recurrent Module} allows to find a higher number of detections, suggesting that the synthesized frames are visually similar to the corresponding real ones, as shown in Figure~\ref{fig:od_cs}.
Through these tests, we verify the capability of the proposed framework to preserve objects and semantic information in the synthesized frames, which is mandatory for employing the proposed method in real-world automotive scenarios.

Finally, we conduct a cross-modality test: we train our model on \textit{DDD17} and \textit{MVSEC} datasets, using as event frame the logarithmic difference of two consecutive frames (\textit{i.e.}~simulated event frames).
Then, we test the network using as input real event frames, without any fine-tuning procedure.
On the \textit{DDD17} dataset we obtain PSNR of $23.396$ and SSIM of $0.779$, and values of $21.935$ and $0.736$ on the \textit{MVSEC} dataset.
These results confirm the ability of the proposed system to deal with both simulated (during train) and real (during test) event data.
Furthermore, it is proved that the logarithmic difference of gray-scale images can be efficiently used in place of real event frames, introducing the possibility to use common RGB dataset and their annotations to simulate the input of an event camera, in the form of event frames.

In Figure~\ref{fig:graphs_metrics}, we plot \textit{L2} and \textit{SSIM} values for each frame within a synthesized sequence, showing the performance drop with respect to the number of generated frames from the last key-frame.
As expected, the contribution of the \textit{Recurrent Module} increases along with the length of the sequence, confirming the effectiveness of the proposed model in the long-sequence generation task.

Our system implementation, tested on a \textit{NVidia 1080Ti}, takes an average time of $47.6 \pm 3.7$ms  to synthesize a single image, reaching a frame rate of about $20$Hz.

\section{\uppercase{Conclusion}}
In this paper, we 
propose a framework able to synthesize color frames, relying on an initial or a periodic set of key-frames and a sequence of event frames. 
The Generative Module produces an intermediate output, while the Recurrent Module refines it, preserving colors and enhancing the temporal coherence.
The method is tested on four public automotive datasets, obtaining state-of-art results.\\
Moreover, semantic segmentation and object detection scores show the possibility to run traditional vision algorithms on synthesized frames, reducing the need of developing new algorithms or collecting new annotated datasets.

\section*{ACKNOWLEDGEMENTS}
This  work  has  been  partially  founded  by  the  project \enquote{Far 2019 -
Metodi per la collaborazione sicura tra operatori umani e sistemi
robotici} (Methods for safe collaboration between human operators and
robotic systems) of the University of Modena and Reggio Emilia.

\bibliographystyle{apalike}
{\small
\bibliography{ms}}

\begin{thebibliography}{}

\bibitem[Andreopoulos et~al., 2018]{andreopoulos2018low}
Andreopoulos, A., Kashyap, H.~J., Nayak, T.~K., Amir, A., and Flickner, M.~D.
  (2018).
\newblock A low power, high throughput, fully event-based stereo system.
\newblock In {\em IEEE International Conference on Computer Vision and Pattern
  Recognition}, pages 7532--7542.

\bibitem[Bardow et~al., 2016]{bardow2016simultaneous}
Bardow, P., Davison, A.~J., and Leutenegger, S. (2016).
\newblock Simultaneous optical flow and intensity estimation from an event
  camera.
\newblock In {\em IEEE International Conference on Computer Vision and Pattern
  Recognition}, pages 884--892.

\bibitem[Binas et~al., 2017]{binas2017ddd17}
Binas, J., Neil, D., Liu, S.-C., and Delbruck, T. (2017).
\newblock Ddd17: End-to-end davis driving dataset.
\newblock {\em Workshop on Machine Learning for Autonomous Vehicles (MLAV) in
  ICML 2017}.

\bibitem[Borghi et~al., 2018]{borghi2018face}
Borghi, G., Fabbri, M., Vezzani, R., Cucchiara, R., et~al. (2018).
\newblock Face-from-depth for head pose estimation on depth images.
\newblock {\em IEEE transactions on pattern analysis and machine intelligence}.

\bibitem[Borghi et~al., 2019]{borghi2019driver}
Borghi, G., Pini, S., Vezzani, R., and Cucchiara, R. (2019).
\newblock Driver face verification with depth maps.
\newblock {\em Sensors}, 19(15):3361.

\bibitem[Brandli et~al., 2014a]{brandli2014240}
Brandli, C., Berner, R., Yang, M., Liu, S.-C., and Delbruck, T. (2014a).
\newblock A 240$\times$ 180 130 db 3 $\mu$s latency global shutter
  spatiotemporal vision sensor.
\newblock {\em IEEE Journal of Solid-State Circuits}, 49(10):2333--2341.

\bibitem[Brandli et~al., 2014b]{brandli2014real}
Brandli, C., Muller, L., and Delbruck, T. (2014b).
\newblock Real-time, high-speed video decompression using a frame-and
  event-based davis sensor.
\newblock In {\em 2014 IEEE International Symposium on Circuits and Systems
  (ISCAS)}, pages 686--689. IEEE.

\bibitem[Cordts et~al., 2016]{cordts2016cityscapes}
Cordts, M., Omran, M., Ramos, S., Rehfeld, T., Enzweiler, M., Benenson, R.,
  Franke, U., Roth, S., and Schiele, B. (2016).
\newblock The cityscapes dataset for semantic urban scene understanding.
\newblock In {\em IEEE International Conference on Computer Vision and Pattern
  Recognition}, pages 3213--3223.

\bibitem[Eigen et~al., 2014]{eigen2014depth}
Eigen, D., Puhrsch, C., and Fergus, R. (2014).
\newblock Depth map prediction from a single image using a multi-scale deep
  network.
\newblock In {\em Neural Information Processing Systems}, pages 2366--2374.

\bibitem[Frigieri et~al., 2017]{frigieri2017fast}
Frigieri, E., Borghi, G., Vezzani, R., and Cucchiara, R. (2017).
\newblock Fast and accurate facial landmark localization in depth images for
  in-car applications.
\newblock In {\em International Conference on Image Analysis and Processing},
  pages 539--549. Springer.

\bibitem[Gallego et~al., 2018a]{gallego2018event}
Gallego, G., Lund, J.~E., Mueggler, E., Rebecq, H., Delbruck, T., and
  Scaramuzza, D. (2018a).
\newblock Event-based, 6-dof camera tracking from photometric depth maps.
\newblock {\em IEEE Transactions on Pattern Analysis and Machine Intelligence},
  40(10):2402--2412.

\bibitem[Gallego et~al., 2018b]{gallego2018unifying}
Gallego, G., Rebecq, H., and Scaramuzza, D. (2018b).
\newblock A unifying contrast maximization framework for event cameras, with
  applications to motion, depth, and optical flow estimation.
\newblock In {\em IEEE International Conference on Computer Vision and Pattern
  Recognition}, volume~1.

\bibitem[Gehrig et~al., 2018]{gehrig2018asynchronous}
Gehrig, D., Rebecq, H., Gallego, G., and Scaramuzza, D. (2018).
\newblock Asynchronous, photometric feature tracking using events and frames.
\newblock In {\em European Conference on Computer Vision}.

\bibitem[Geiger et~al., 2013]{Geiger2013IJRR}
Geiger, A., Lenz, P., Stiller, C., and Urtasun, R. (2013).
\newblock Vision meets robotics: The kitti dataset.
\newblock {\em International Journal of Robotics Research (IJRR)}.

\bibitem[Geiger et~al., 2012]{Geiger2012CVPR}
Geiger, A., Lenz, P., and Urtasun, R. (2012).
\newblock Are we ready for autonomous driving? the kitti vision benchmark
  suite.
\newblock In {\em IEEE International Conference on Computer Vision and Pattern
  Recognition}.

\bibitem[Goodfellow et~al., 2014]{goodfellow2014generative}
Goodfellow, I., Pouget-Abadie, J., Mirza, M., Xu, B., Warde-Farley, D., Ozair,
  S., Courville, A., and Bengio, Y. (2014).
\newblock Generative adversarial nets.
\newblock In {\em Neural Information Processing Systems}, pages 2672--2680.

\bibitem[Isola et~al., 2017]{isola2017image}
Isola, P., Zhu, J.-Y., Zhou, T., and Efros, A.~A. (2017).
\newblock Image-to-image translation with conditional adversarial networks.
\newblock In {\em IEEE International Conference on Computer Vision and Pattern
  Recognition}.

\bibitem[Kim et~al., 2014]{KimHBID14}
Kim, H., Handa, A., Benosman, R., Ieng, S., and Davison, A.~J. (2014).
\newblock Simultaneous mosaicing and tracking with an event camera.
\newblock In {\em British Machine Vision Conference}.

\bibitem[Kingma and Ba, 2014]{KingmaB14}
Kingma, D.~P. and Ba, J. (2014).
\newblock Adam: {A} method for stochastic optimization.
\newblock {\em CoRR}, abs/1412.6980.

\bibitem[Lagorce et~al., 2017]{lagorce2017hots}
Lagorce, X., Orchard, G., Galluppi, F., Shi, B.~E., and Benosman, R.~B. (2017).
\newblock Hots: a hierarchy of event-based time-surfaces for pattern
  recognition.
\newblock {\em IEEE Transactions on Pattern Analysis and Machine Intelligence},
  39(7):1346--1359.

\bibitem[Lichtsteiner et~al., 2006]{lichtsteiner2008128}
Lichtsteiner, P., Posch, C., and Delbruck, T. (2006).
\newblock A 128 x 128 120db 30mw asynchronous vision sensor that responds to
  relative intensity change.
\newblock In {\em Solid-State Circuits Conference, 2006. ISSCC 2006. Digest of
  Technical Papers. IEEE International}, pages 2060--2069. IEEE.

\bibitem[Lin et~al., 2014]{lin2014microsoftcoco}
Lin, T.-Y., Maire, M., Belongie, S., Hays, J., Perona, P., Ramanan, D.,
  Doll{\'a}r, P., and Zitnick, C.~L. (2014).
\newblock Microsoft {COCO}: Common objects in context.
\newblock In {\em European Conference on Computer Vision}. Springer.

\bibitem[Lungu et~al., 2017]{lungu2017live}
Lungu, I.-A., Corradi, F., and Delbr{\"u}ck, T. (2017).
\newblock Live demonstration: Convolutional neural network driven by dynamic
  vision sensor playing roshambo.
\newblock In {\em IEEE International Symposium on Circuits and Systems
  (ISCAS)}, pages 1--1. IEEE.

\bibitem[Maqueda et~al., 2018]{maqueda2018event}
Maqueda, A.~I., Loquercio, A., Gallego, G., Garc{\i}a, N., and Scaramuzza, D.
  (2018).
\newblock Event-based vision meets deep learning on steering prediction for
  self-driving cars.
\newblock In {\em IEEE International Conference on Computer Vision and Pattern
  Recognition}, pages 5419--5427.

\bibitem[Mirza and Osindero, 2014]{mirza2014conditional}
Mirza, M. and Osindero, S. (2014).
\newblock Conditional generative adversarial nets.
\newblock {\em arXiv preprint arXiv:1411.1784}.

\bibitem[Mitrokhin et~al., 2018]{mitrokhin2018event}
Mitrokhin, A., Fermuller, C., Parameshwara, C., and Aloimonos, Y. (2018).
\newblock Event-based moving object detection and tracking.
\newblock {\em arXiv preprint arXiv:1803.04523}.

\bibitem[Munda et~al., 2018]{munda2018real}
Munda, G., Reinbacher, C., and Pock, T. (2018).
\newblock Real-time intensity-image reconstruction for event cameras using
  manifold regularisation.
\newblock {\em International Journal of Computer Vision}, 126(12):1381--1393.

\bibitem[Pini et~al., 2019]{pini2019video}
Pini, S., Borghi, G., Vezzani, R., and Cucchiara, R. (2019).
\newblock Video synthesis from intensity and event frames.
\newblock In {\em International Conference on Image Analysis and Processing},
  pages 313--323. Springer.

\bibitem[Ramesh et~al., 2018]{rameshlong}
Ramesh, B., Zhang, S., Lee, Z.~W., Gao, Z., Orchard, G., and Xiang, C. (2018).
\newblock Long-term object tracking with a moving event camera.
\newblock In {\em British Machine Vision Conference}.

\bibitem[Rebecq et~al., 2016]{rebecq2016emvs}
Rebecq, H., Gallego, G., and Scaramuzza, D. (2016).
\newblock Emvs: Event-based multi-view stereo.
\newblock In {\em British Machine Vision Conference}.

\bibitem[Rebecq et~al., 2019]{rebecq2019events}
Rebecq, H., Ranftl, R., Koltun, V., and Scaramuzza, D. (2019).
\newblock Events-to-video: Bringing modern computer vision to event cameras.
\newblock In {\em Proceedings of the IEEE Conference on Computer Vision and
  Pattern Recognition}, pages 3857--3866.

\bibitem[Redmon and Farhadi, 2018]{yolov3}
Redmon, J. and Farhadi, A. (2018).
\newblock {YOLO}v3: An incremental improvement.
\newblock {\em arXiv preprint arXiv:1804.02767}.

\bibitem[Reinbacher et~al., 2016]{reinbacher2016real}
Reinbacher, C., Graber, G., and Pock, T. (2016).
\newblock Real-time intensity-image reconstruction for event cameras using
  manifold regularisation.
\newblock In {\em British Machine Vision Conference}.

\bibitem[Ronneberger et~al., 2015]{ronneberger2015u}
Ronneberger, O., Fischer, P., and Brox, T. (2015).
\newblock U-net: Convolutional networks for biomedical image segmentation.
\newblock In {\em International Conference on Medical image computing and
  computer-assisted intervention}, pages 234--241. Springer.

\bibitem[Rota~Bul\`o et~al., 2018]{rotabulo2017place}
Rota~Bul\`o, S., Porzi, L., and Kontschieder, P. (2018).
\newblock In-place activated batchnorm for memory-optimized training of dnns.
\newblock In {\em Proceedings of the IEEE Conference on Computer Vision and
  Pattern Recognition}.

\bibitem[Salimans et~al., 2016]{salimans2016improved}
Salimans, T., Goodfellow, I., Zaremba, W., Cheung, V., Radford, A., and Chen,
  X. (2016).
\newblock Improved techniques for training gans.
\newblock In {\em Neural Information Processing Systems}, pages 2234--2242.

\bibitem[Scheerlinck et~al., 2018]{scheerlinck2018continuous}
Scheerlinck, C., Barnes, N., and Mahony, R. (2018).
\newblock Continuous-time intensity estimation using event cameras.
\newblock {\em Asian Conf. Comput. Vis. (ACCV)}.

\bibitem[Uhrig et~al., 2017]{Uhrig2017THREEDV}
Uhrig, J., Schneider, N., Schneider, L., Franke, U., Brox, T., and Geiger, A.
  (2017).
\newblock Sparsity invariant cnns.
\newblock In {\em International Conference on 3D Vision (3DV)}.

\bibitem[Wang et~al., 2004]{wang2004image}
Wang, Z., Bovik, A.~C., Sheikh, H.~R., and Simoncelli, E.~P. (2004).
\newblock Image quality assessment: from error visibility to structural
  similarity.
\newblock {\em IEEE transactions on image processing}, 13(4):600--612.

\bibitem[Xingjian et~al., 2015]{xingjian2015convolutional}
Xingjian, S., Chen, Z., Wang, H., Yeung, D.-Y., Wong, W.-K., and Woo, W.-c.
  (2015).
\newblock Convolutional lstm network: A machine learning approach for
  precipitation nowcasting.
\newblock In {\em Neural Information Processing Systems}, pages 802--810.

\bibitem[Zhang et~al., 2018]{zhang2018perceptual}
Zhang, R., Isola, P., Efros, A.~A., Shechtman, E., and Wang, O. (2018).
\newblock The unreasonable effectiveness of deep features as a perceptual
  metric.
\newblock In {\em CVPR}.

\bibitem[Zhou et~al., 2018]{zhou2018semi}
Zhou, Y., Gallego, G., Rebecq, H., Kneip, L., Li, H., and Scaramuzza, D.
  (2018).
\newblock Semi-dense 3d reconstruction with a stereo event camera.
\newblock In {\em European Conference on Computer Vision}.

\bibitem[Zhu et~al., 2018a]{zhu2018multi}
Zhu, A.~Z., Thakur, D., Ozaslan, T., Pfrommer, B., Kumar, V., and Daniilidis,
  K. (2018a).
\newblock The multi vehicle stereo event camera dataset: An event camera
  dataset for 3d perception.
\newblock {\em IEEE Robotics and Automation Letters}, 3(3):2032--2039.

\bibitem[Zhu et~al., 2018b]{zhu2018multivehicle}
Zhu, A.~Z., Thakur, D., {\"O}zaslan, T., Pfrommer, B., Kumar, V., and
  Daniilidis, K. (2018b).
\newblock The multivehicle stereo event camera dataset: An event camera dataset
  for 3d perception.
\newblock {\em IEEE Robotics and Automation Letters}, 3(3):2032--2039.

\end{thebibliography}


\twocolumn[  
    \begin{@twocolumnfalse}
         \section*{\uppercase{Supplementary Material}}
    \end{@twocolumnfalse}
]
In this supplementary material, we present some additional details, in form of mathematical formulations and output images, not included in the original paper due to page limitation. \\
Specifically, we report the mathematical formulation of the pixel-wise metrics, in Section~\ref{formula}. Moreover, in Section~\ref{visual_results}, we include additional generated images along with outputs of the semantic segmentation and object detection algorithms.

\subsection*{Pixel-wise metrics formulas} \label{formula}
Metrics reported in Table 1 and Table 2 in the paper are defined as follows.
Given a ground truth image $I$ and a synthesized image $\hat I$, we refer with $\hat y \in \hat I$ as the element of the generated image at the same location of $y \in I$.

\begin{itemize}

    \item  $L_1$ and $L_2$ norms:

\begin{equation}
    L_1 = \frac{1}{|I|} \, \sum_{y \in I} | \, y - \hat{y} \, |
\end{equation}

\begin{equation}
    L_2 = \frac{1}{|I|} \, \sum_{y \in I} ( \, y - \hat{y} \, )^2
\end{equation}




    \item Root Mean Squared Error (RMSE):

\begin{equation}
    \text{Lin} = \sqrt{ \frac{1}{|I|} \, \sum_{y \in I} ( \, y - \hat{y} \, )^2 }
\end{equation}

\begin{equation}
    \text{Log} =  \sqrt{ \frac{1}{|I|} \, \sum_{y \in I} ( \, \log y - \log \hat{y} \, )^2 }
\end{equation}

\begin{equation}
    \text{Scl} =  \frac{1}{|I|} \sum_{y \in I} d^2 - \frac{1}{|I|^2} \big( \sum_{y \in I} d \big)^2 
\end{equation}
$$d = \log y - \log \hat{y}$$

    \item Thresholds:
    
\begin{equation}
    \delta_i = \frac{1}{|I|} \, |J_i| \, , \quad i \in \{1, 2, 3\}
\end{equation}
$$J_i = \{y \in I \, | \, max\big(\frac{y}{\hat{y}}, \frac{\hat{y}}{y}\big) < 1.25^i\}$$

    \item Indexes:

\begin{equation}
    \text{PSNR} = 10 \cdot  \log_{10} \big(\frac{m}{L_2}\big)
\end{equation}
where $m$ is the maximum possible value of $I$ and $\hat{I}$. In our experiments, $m=1$.

\begin{equation}
	\text{SSIM}(p,q) = \frac{(2\mu_p\mu_q + c_1)(2\sigma_{pq} + c_2)}{(\mu_p^2 + \mu_q^2 + c_1)(\sigma_p^2 + \sigma_q^2 + c_2)}
\end{equation}

Given two windows $p \in I$, $q \in \hat{I}$ of equal size $11 \times 11$, $\mu_{p,q}$, $\sigma_{p,q}$ are the mean and variance of $p$ and $q$, $\sigma_{pq}$ is the covariance of $p$, $q$. \\
Finally, $c_{1,2}$ are defined as $c_1 = (0.01 \cdot L)^2$ and $c_2 = (0.03 \cdot L)^2$ where $L$ is the dynamic range (\ie the difference between the maximum and the minimum theoretical value) of $I$ and $\hat{I}$. In our experiments, $L=1$.\\
For further details about these metrics, see the original papers~\cite{eigen2014depth,wang2004image}.

\end{itemize}


\subsection*{Perceptual evaluation}
As stated in the paper, in order to assess the perceived quality of the generated images we exploit LPIPS metric that resembles the human visual perception. 
In particular, LPIPS~\cite{zhang2018perceptual} measures the similarity between image patches as the distance between the activations of a deep neural network.
Further information are available on the original work and on the project webpage\footnote{\url{https://github.com/richzhang/PerceptualSimilarity}}.

\subsection*{Sample visual results} \label{visual_results}
\begin{figure*}[th]
\captionsetup[subfigure]{labelformat=empty}
\centering
\resizebox{0.95\textwidth}{!}{
    \centering
    \begin{subfigure}[t]{0.19\textwidth}
        \centering
        \caption{{Ground Truth}}
        \includegraphics[width=0.98\linewidth]{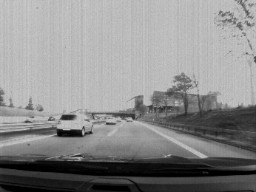}\vspace{2px}
        \includegraphics[width=0.98\linewidth]{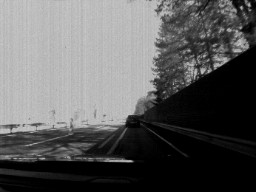}\vspace{2px}
        \includegraphics[width=0.98\linewidth]{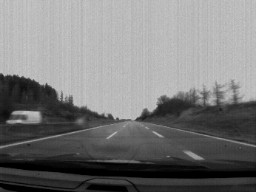}\vspace{2px}
        \includegraphics[width=0.98\linewidth]{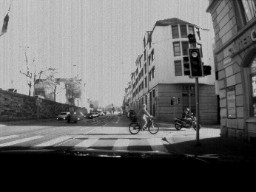}\vspace{2px}
        \includegraphics[width=0.98\linewidth]{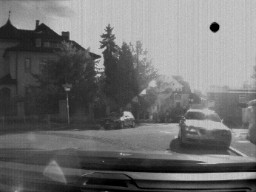}\vspace{2px}
        \includegraphics[width=0.98\linewidth]{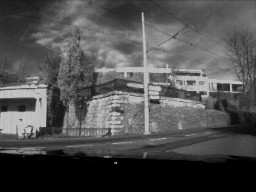}\vspace{2px}
        \includegraphics[width=0.98\linewidth]{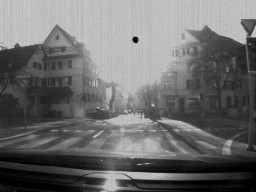}\vspace{2px}
    \end{subfigure}
    \centering
    \begin{subfigure}[t]{0.19\textwidth}
        \centering
        \caption{{Event Frames}}
        \includegraphics[width=0.98\linewidth]{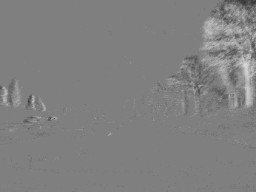}\vspace{2px}
        \includegraphics[width=0.98\linewidth]{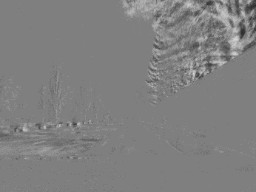}\vspace{2px}
        \includegraphics[width=0.98\linewidth]{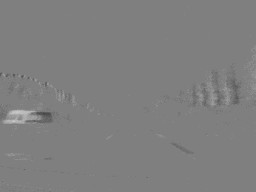}\vspace{2px}
        \includegraphics[width=0.98\linewidth]{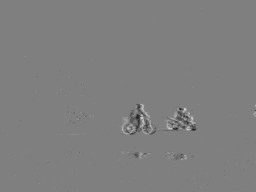}\vspace{2px}
        \includegraphics[width=0.98\linewidth]{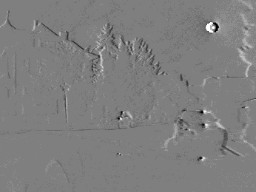}\vspace{2px}
        \includegraphics[width=0.98\linewidth]{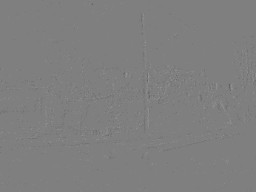}\vspace{2px}
        \includegraphics[width=0.98\linewidth]{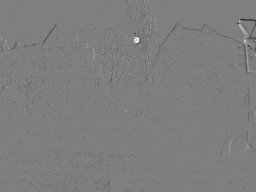}\vspace{2px}
    \end{subfigure}
    \centering
    \begin{subfigure}[t]{0.19\textwidth}
        \centering
        \caption{{Munda \textit{et al.} }}
        \includegraphics[width=0.98\linewidth]{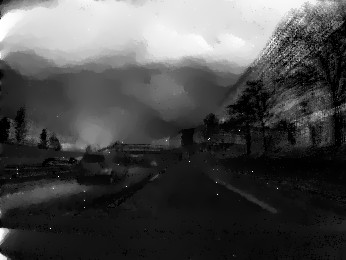}\vspace{2px}
        \includegraphics[width=0.98\linewidth]{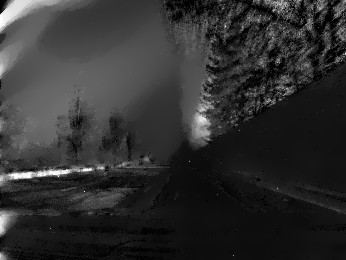}\vspace{2px}
        \includegraphics[width=0.98\linewidth]{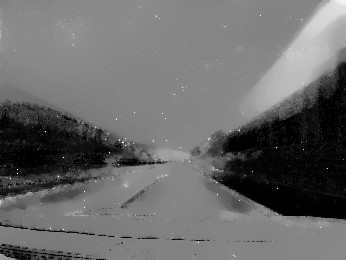}\vspace{2px}
        \includegraphics[width=0.98\linewidth]{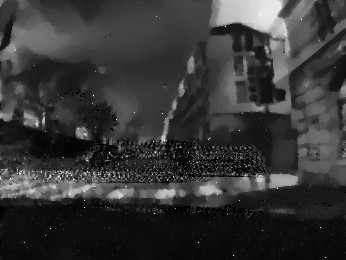}\vspace{2px}
        \includegraphics[width=0.98\linewidth]{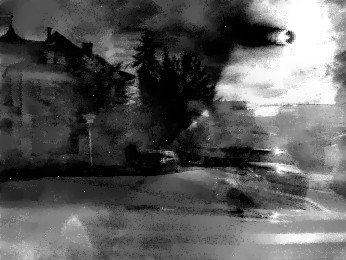}\vspace{2px}
        \includegraphics[width=0.98\linewidth]{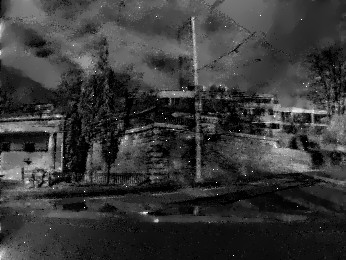}\vspace{2px}
        \includegraphics[width=0.98\linewidth]{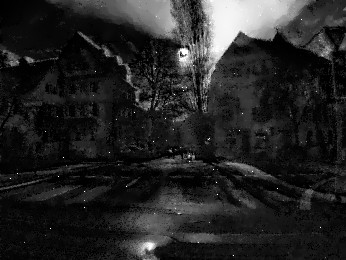}\vspace{2px}
    \end{subfigure}
    \begin{subfigure}[t]{0.19\textwidth}
        \centering
        \caption{{Scheerlinck \textit{et al.} }}
        \includegraphics[width=0.98\linewidth]{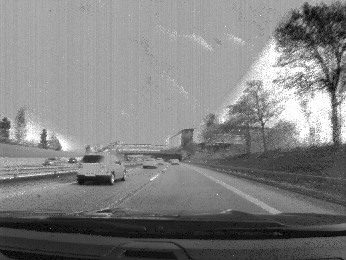}\vspace{2px}
        \includegraphics[width=0.98\linewidth]{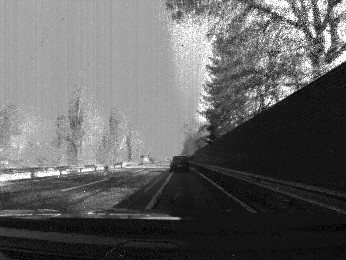}\vspace{2px}
        \includegraphics[width=0.98\linewidth]{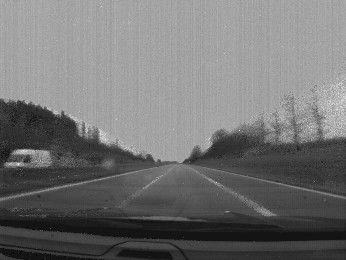}\vspace{2px}
        \includegraphics[width=0.98\linewidth]{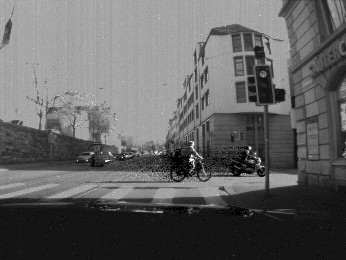}\vspace{2px}
        \includegraphics[width=0.98\linewidth]{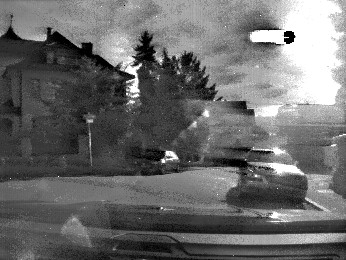}\vspace{2px}
        \includegraphics[width=0.98\linewidth]{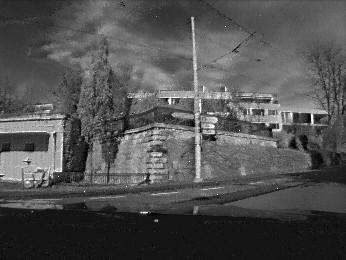}\vspace{2px}
        \includegraphics[width=0.98\linewidth]{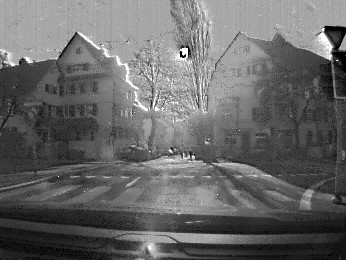}\vspace{2px}
    \end{subfigure}
        \begin{subfigure}[t]{0.19\textwidth}
        \centering
        \caption{{Ours}}
        \includegraphics[width=0.98\linewidth]{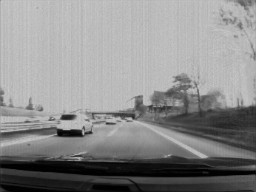}\vspace{2px}
        \includegraphics[width=0.98\linewidth]{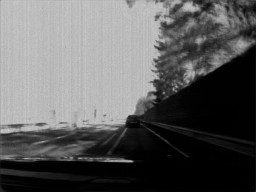}\vspace{2px}
        \includegraphics[width=0.98\linewidth]{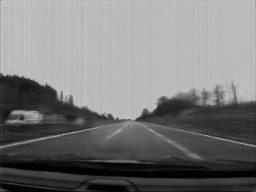}\vspace{2px}
        \includegraphics[width=0.98\linewidth]{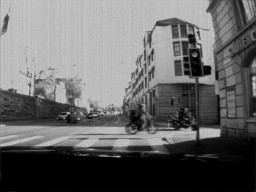}\vspace{2px}
        \includegraphics[width=0.98\linewidth]{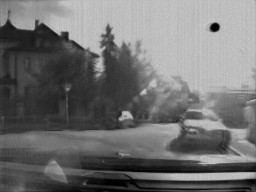}\vspace{2px}
        \includegraphics[width=0.98\linewidth]{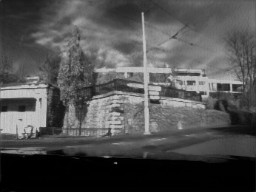}\vspace{2px}
        \includegraphics[width=0.98\linewidth]{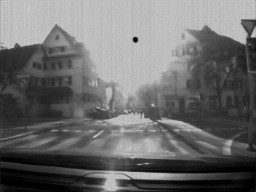}\vspace{2px}
    \end{subfigure}
}
    \caption{Output of the proposed method compared with~\cite{munda2018real,scheerlinck2018continuous} on DDD17 dataset~\cite{binas2017ddd17}.}
    \label{fig:gen_ddd}
\end{figure*}

\begin{figure*}[th]
\captionsetup[subfigure]{labelformat=empty}
\centering
\resizebox{0.95\textwidth}{!}{
    \centering
    \begin{subfigure}[t]{0.19\textwidth}
        \centering
        \caption{{Ground Truth}}
        \includegraphics[width=0.98\linewidth]{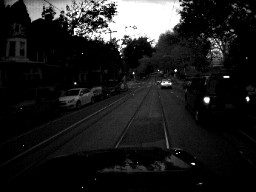}\vspace{2px}
        \includegraphics[width=0.98\linewidth]{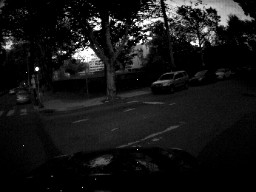}\vspace{2px}
        \includegraphics[width=0.98\linewidth]{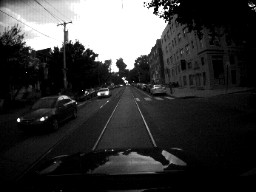}\vspace{2px}
        \includegraphics[width=0.98\linewidth]{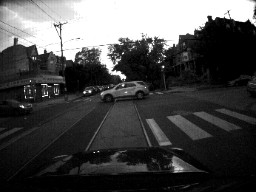}\vspace{2px}
        \includegraphics[width=0.98\linewidth]{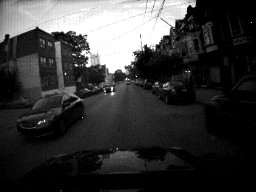}\vspace{2px}
    \end{subfigure}
    \centering
    \begin{subfigure}[t]{0.19\textwidth}
        \centering
        \caption{{Event Frames}}
        \includegraphics[width=0.98\linewidth]{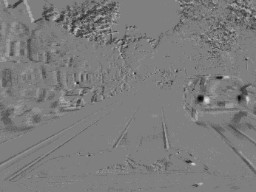}\vspace{2px}
        \includegraphics[width=0.98\linewidth]{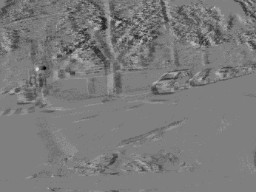}\vspace{2px}
        \includegraphics[width=0.98\linewidth]{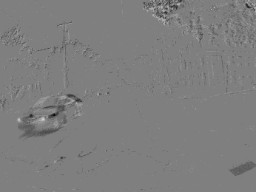}\vspace{2px}
        \includegraphics[width=0.98\linewidth]{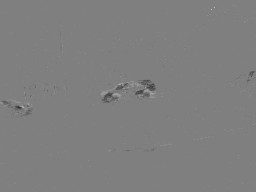}\vspace{2px}
        \includegraphics[width=0.98\linewidth]{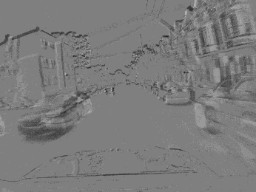}\vspace{2px}
    \end{subfigure}
    \centering
    \begin{subfigure}[t]{0.19\textwidth}
        \centering
        \caption{{Munda \textit{et al.}}}
        \includegraphics[width=0.98\linewidth]{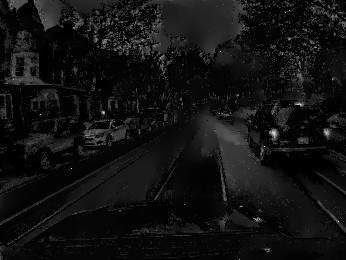}\vspace{2px}
        \includegraphics[width=0.98\linewidth]{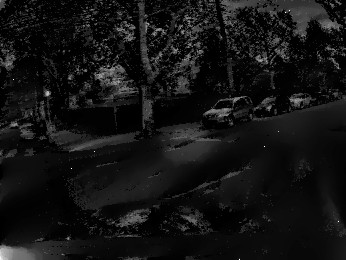}\vspace{2px}
        \includegraphics[width=0.98\linewidth]{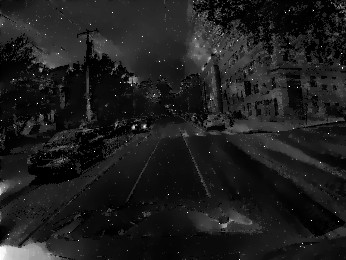}\vspace{2px}
        \includegraphics[width=0.98\linewidth]{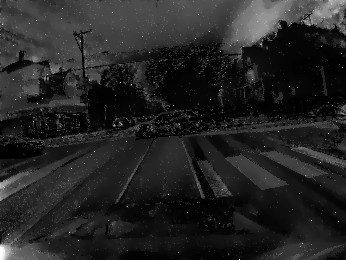}\vspace{2px}
        \includegraphics[width=0.98\linewidth]{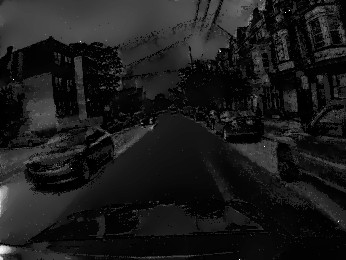}\vspace{2px}
    \end{subfigure}
    \begin{subfigure}[t]{0.19\textwidth}
        \centering
        \caption{{Scheerlinck \textit{et al.}}}
        \includegraphics[width=0.98\linewidth]{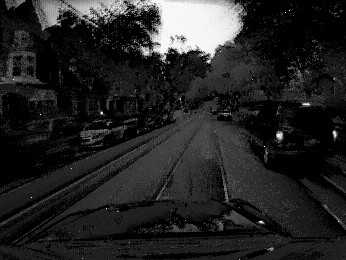}\vspace{2px}
        \includegraphics[width=0.98\linewidth]{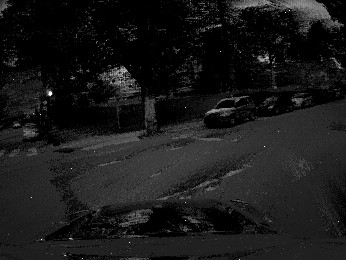}\vspace{2px}
        \includegraphics[width=0.98\linewidth]{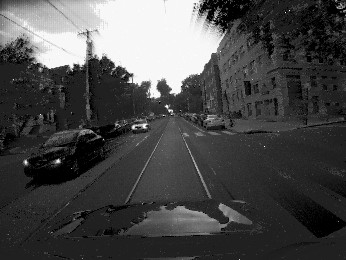}\vspace{2px}
        \includegraphics[width=0.98\linewidth]{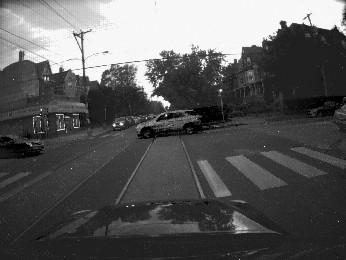}\vspace{2px}
        \includegraphics[width=0.98\linewidth]{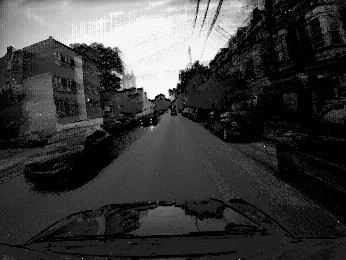}\vspace{2px}
    \end{subfigure}
        \begin{subfigure}[t]{0.19\textwidth}
        \centering
        \caption{{Ours}}
        \includegraphics[width=0.98\linewidth]{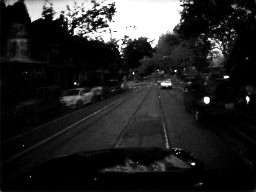}\vspace{2px}
        \includegraphics[width=0.98\linewidth]{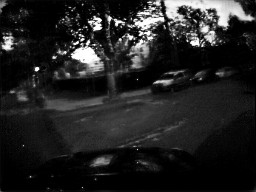}\vspace{2px}
        \includegraphics[width=0.98\linewidth]{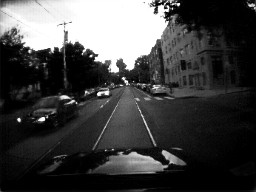}\vspace{2px}
        \includegraphics[width=0.98\linewidth]{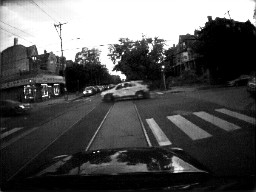}\vspace{2px}
        \includegraphics[width=0.98\linewidth]{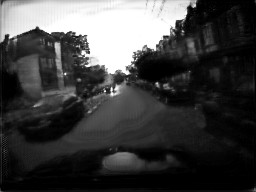}\vspace{2px}
    \end{subfigure}
}
    \caption{Output of the proposed method compared with~\cite{munda2018real,scheerlinck2018continuous} on MVSEC dataset~\cite{zhu2018multi}. 
    Images are enhanced on the brightness only for a better visualization.
    }
    \label{fig:gen_mvsec}
\end{figure*}

\begin{figure*}[th]
\captionsetup[subfigure]{labelformat=empty}
\centering
\resizebox{0.65\textwidth}{!}{
    \centering
    \begin{subfigure}[t]{0.19\textwidth}
        \centering
        \caption{{\small ~\\Ground truth}}
        \includegraphics[width=0.98\linewidth]{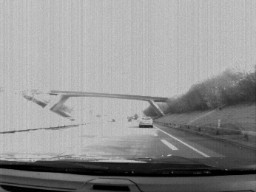}\vspace{2px}
        \includegraphics[width=0.98\linewidth]{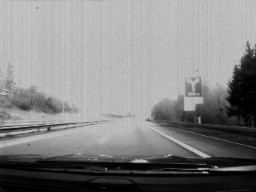}\vspace{2px}
        \includegraphics[width=0.98\linewidth]{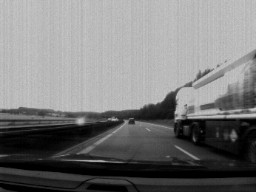}\vspace{2px}
        \includegraphics[width=0.98\linewidth]{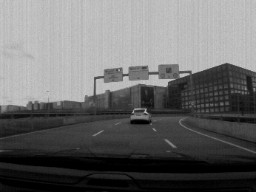}\vspace{2px}
        \includegraphics[width=0.98\linewidth]{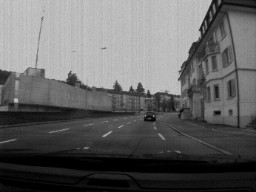}\vspace{2px}
        \includegraphics[width=0.98\linewidth]{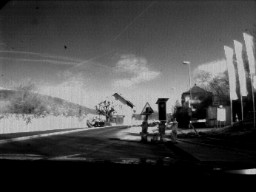}\vspace{2px}
        \includegraphics[width=0.98\linewidth]{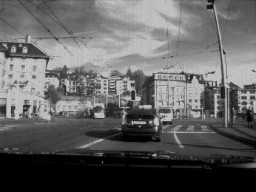}\vspace{2px}
    \end{subfigure}
    \centering
    \begin{subfigure}[t]{0.19\textwidth}
        \centering
        \caption{{\small \centerline{Ours} \\ \centerline{D.I. as Event Frames}}}
        \includegraphics[width=0.98\linewidth]{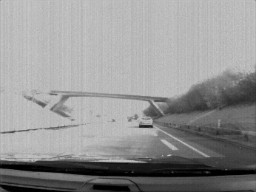}\vspace{2px}
        \includegraphics[width=0.98\linewidth]{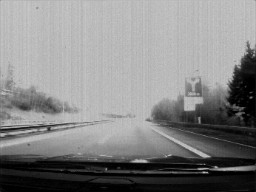}\vspace{2px}
        \includegraphics[width=0.98\linewidth]{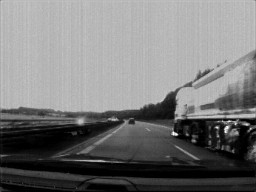}\vspace{2px}
        \includegraphics[width=0.98\linewidth]{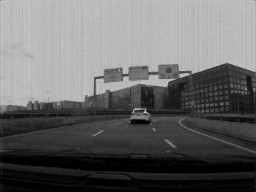}\vspace{2px}
        \includegraphics[width=0.98\linewidth]{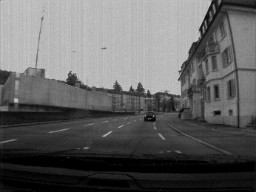}\vspace{2px}
        \includegraphics[width=0.98\linewidth]{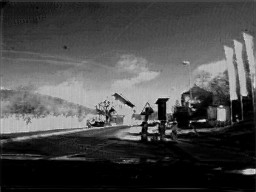}\vspace{2px}
        \includegraphics[width=0.98\linewidth]{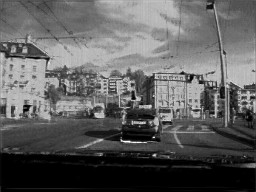}\vspace{2px}
    \end{subfigure}
    \centering
    \begin{subfigure}[t]{0.19\textwidth}
        \centering
        \caption{{\small \centerline{Ours}\\ \centerline{Real Event Frames}}}
        \includegraphics[width=0.98\linewidth]{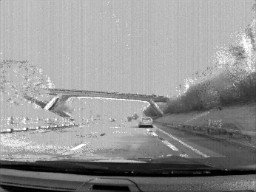}\vspace{2px}
        \includegraphics[width=0.98\linewidth]{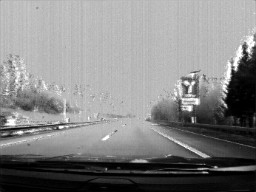}\vspace{2px}
        \includegraphics[width=0.98\linewidth]{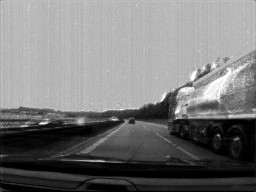}\vspace{2px}
        \includegraphics[width=0.98\linewidth]{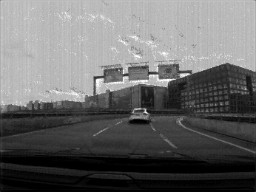}\vspace{2px}
        \includegraphics[width=0.98\linewidth]{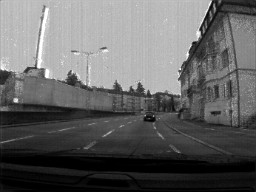}\vspace{2px}
        \includegraphics[width=0.98\linewidth]{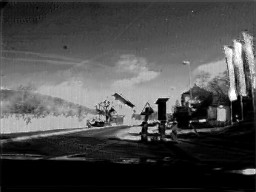}\vspace{2px}
        \includegraphics[width=0.98\linewidth]{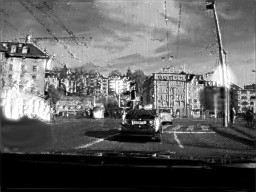}\vspace{2px}
    \end{subfigure}
}
    \caption{Output samples related to the cross-modality test (see Sect.~$5.3$) on DDD17 dataset~\cite{binas2017ddd17}. 
    Given the proposed model trained on event frames obtained as difference on images, in the second and third columns we report generated frames using simulated and real event frames as input, respectively.
    Here, D.I. stands for Difference of Images (see Sect.~$3.3$ in the paper).}
    \label{fig:cross_ddd}
\end{figure*}

\begin{figure*}[th]
\captionsetup[subfigure]{labelformat=empty}
\centering
\resizebox{0.63\textwidth}{!}{
    \centering
    \begin{subfigure}[t]{0.19\textwidth}
        \centering
        \caption{{\small ~\\Ground truth}}
        \includegraphics[width=0.98\linewidth]{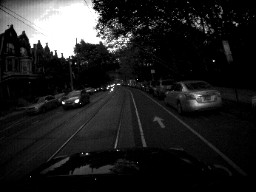}\vspace{2px}
        \includegraphics[width=0.98\linewidth]{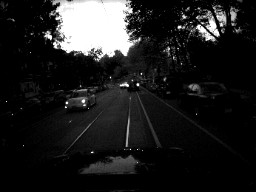}\vspace{2px}
        \includegraphics[width=0.98\linewidth]{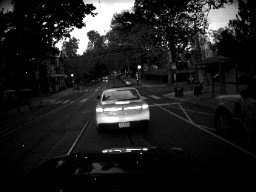}\vspace{2px}
        \includegraphics[width=0.98\linewidth]{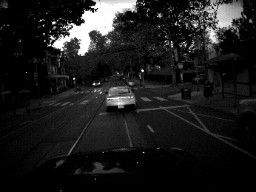}\vspace{2px}
        \includegraphics[width=0.98\linewidth]{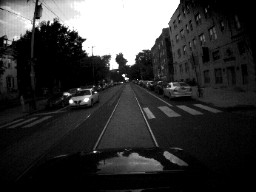}\vspace{2px}
        \includegraphics[width=0.98\linewidth]{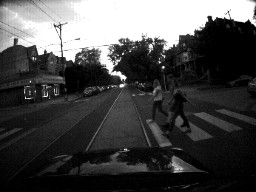}\vspace{2px}
        \includegraphics[width=0.98\linewidth]{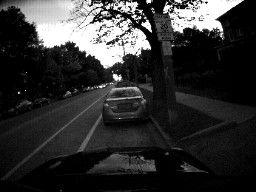}\vspace{2px}
    \end{subfigure}
    \centering
    \begin{subfigure}[t]{0.19\textwidth}
        \centering
        \caption{{\small \centerline{Ours} \\ \centerline{D.I. as Event Frames}}}
        \includegraphics[width=0.98\linewidth]{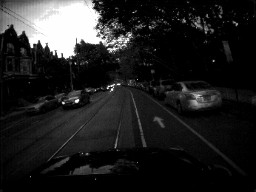}\vspace{2px}
        \includegraphics[width=0.98\linewidth]{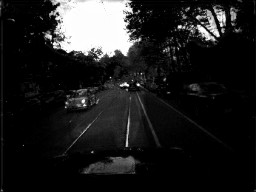}\vspace{2px}
        \includegraphics[width=0.98\linewidth]{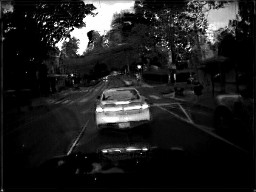}\vspace{2px}
        \includegraphics[width=0.98\linewidth]{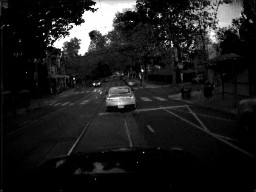}\vspace{2px}
        \includegraphics[width=0.98\linewidth]{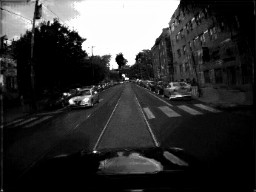}\vspace{2px}
        \includegraphics[width=0.98\linewidth]{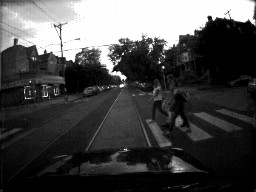}\vspace{2px}
        \includegraphics[width=0.98\linewidth]{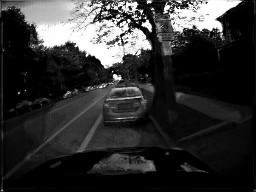}\vspace{2px}
    \end{subfigure}
    \centering
    \begin{subfigure}[t]{0.19\textwidth}
        \centering
        \caption{{\small \centerline{Ours}\\ \centerline{Real Event Frames}}}
        \includegraphics[width=0.98\linewidth]{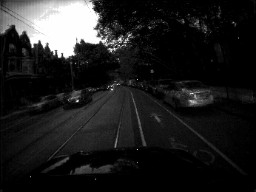}\vspace{2px}
        \includegraphics[width=0.98\linewidth]{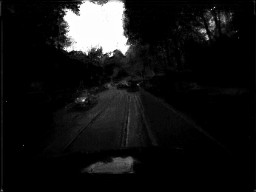}\vspace{2px}
        \includegraphics[width=0.98\linewidth]{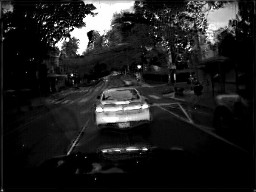}\vspace{2px}
        \includegraphics[width=0.98\linewidth]{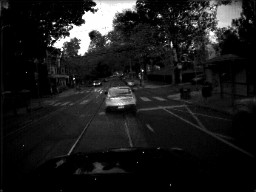}\vspace{2px}
        \includegraphics[width=0.98\linewidth]{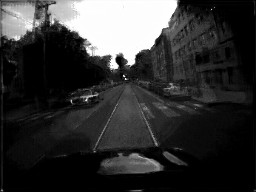}\vspace{2px}
        \includegraphics[width=0.98\linewidth]{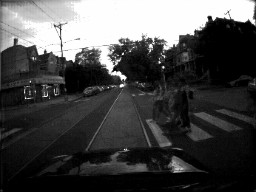}\vspace{2px}
        \includegraphics[width=0.98\linewidth]{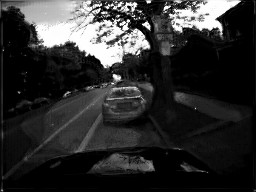}\vspace{2px}
    \end{subfigure}
}
    \caption{Output samples related to the cross-modality test (see Sect.~$5.3$) on MVSEC dataset~\cite{zhu2018multi}. 
    Given the proposed model trained on event frames obtained as difference on images, in the second and third columns we report generated frames using simulated and real event frames as input, respectively.
    Here, D.I. stands for Difference of Images (see Sect.~$3.3$ in the paper).
    Images are enhanced on the brightness only for a better visualization.}
    \label{fig:cross_mvsec}
\end{figure*}

\begin{figure*}[th]
\captionsetup[subfigure]{labelformat=empty}
\centering
    \centering
    \begin{subfigure}[t]{0.49\textwidth}
        \centering
        \caption{{Ground Truth}}
        \includegraphics[width=0.48\linewidth]{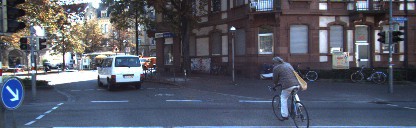}
        \includegraphics[width=0.48\linewidth]{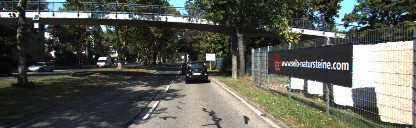}
        \includegraphics[width=0.48\linewidth]{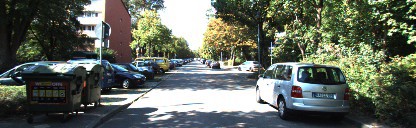}
        \includegraphics[width=0.48\linewidth]{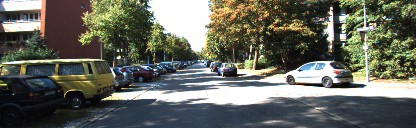}
        \includegraphics[width=0.48\linewidth]{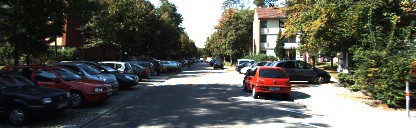}
        \includegraphics[width=0.48\linewidth]{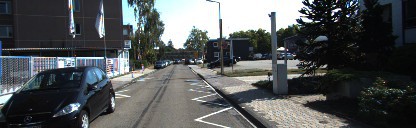}
        \includegraphics[width=0.48\linewidth]{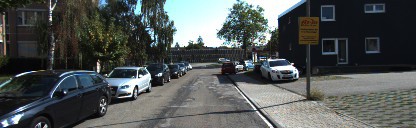}
        \includegraphics[width=0.48\linewidth]{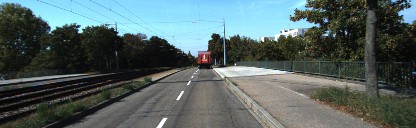}
        \includegraphics[width=0.48\linewidth]{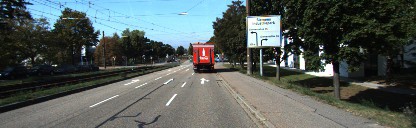}
        \includegraphics[width=0.48\linewidth]{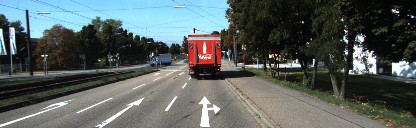}
        \includegraphics[width=0.48\linewidth]{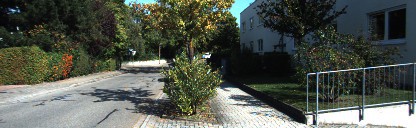}
        \includegraphics[width=0.48\linewidth]{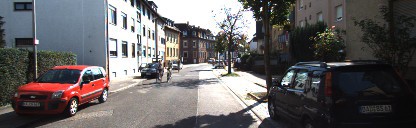}
        \includegraphics[width=0.48\linewidth]{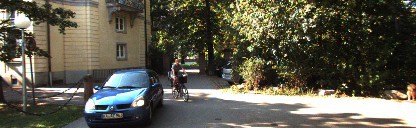}
        \includegraphics[width=0.48\linewidth]{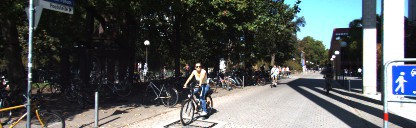}
        \includegraphics[width=0.48\linewidth]{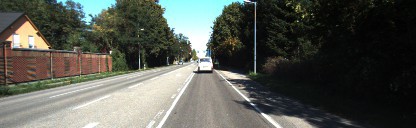}
        \includegraphics[width=0.48\linewidth]{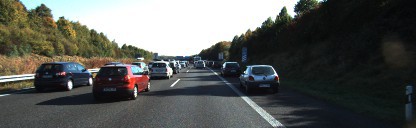}
        \includegraphics[width=0.48\linewidth]{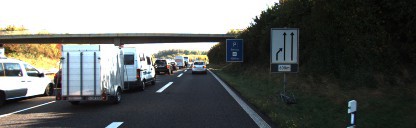}
        \includegraphics[width=0.48\linewidth]{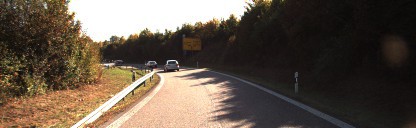}
    \end{subfigure}
    \centering
    \begin{subfigure}[t]{0.49\textwidth}
        \centering
        \caption{{Ours}}
        \includegraphics[width=0.48\linewidth]{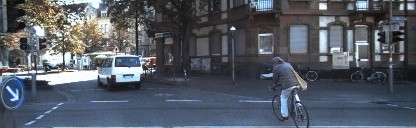}
        \includegraphics[width=0.48\linewidth]{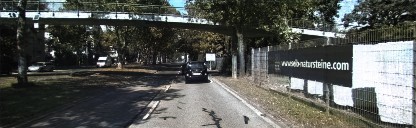}
        \includegraphics[width=0.48\linewidth]{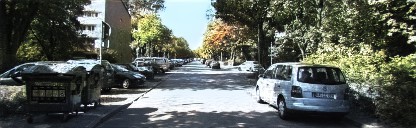}
        \includegraphics[width=0.48\linewidth]{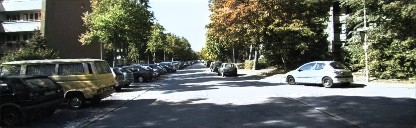}
        \includegraphics[width=0.48\linewidth]{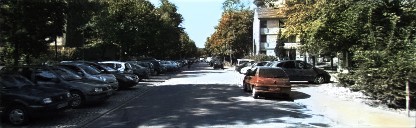}
        \includegraphics[width=0.48\linewidth]{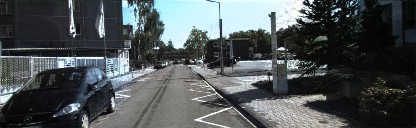}
        \includegraphics[width=0.48\linewidth]{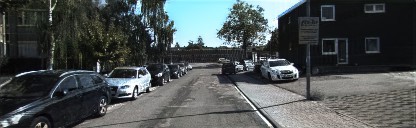}
        \includegraphics[width=0.48\linewidth]{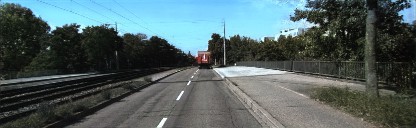}
        \includegraphics[width=0.48\linewidth]{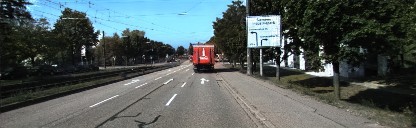}
        \includegraphics[width=0.48\linewidth]{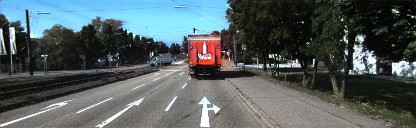}
        \includegraphics[width=0.48\linewidth]{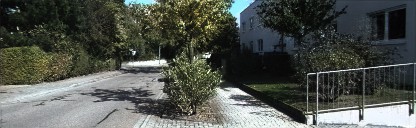}
        \includegraphics[width=0.48\linewidth]{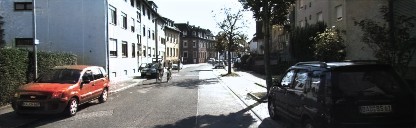}
        \includegraphics[width=0.48\linewidth]{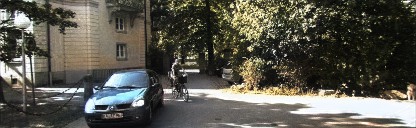}
        \includegraphics[width=0.48\linewidth]{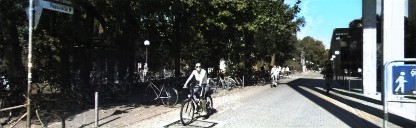}
        \includegraphics[width=0.48\linewidth]{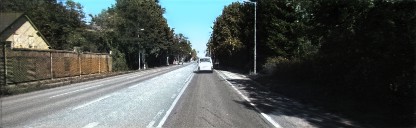}
        \includegraphics[width=0.48\linewidth]{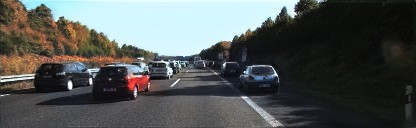}
        \includegraphics[width=0.48\linewidth]{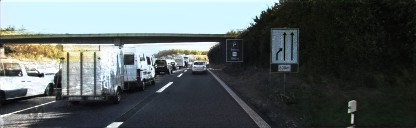}
        \includegraphics[width=0.48\linewidth]{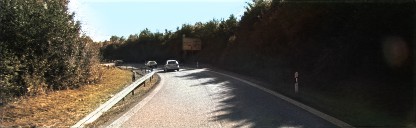}
    \end{subfigure}
    \caption{Output of the proposed algorithm obtained on the \textit{Kitti} dataset~\cite{Geiger2013IJRR}.}
    \label{fig:samples_kitti}
\end{figure*}

\begin{figure*}[th]
\captionsetup[subfigure]{labelformat=empty}
\centering
    \centering
    \begin{subfigure}[t]{0.49\textwidth}
        \centering
        \caption{{Ground Truth}}
        \includegraphics[width=0.48\linewidth]{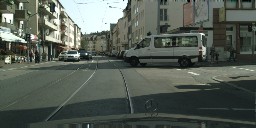}
        \includegraphics[width=0.48\linewidth]{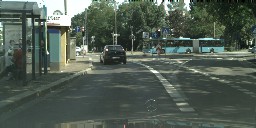}
        \includegraphics[width=0.48\linewidth]{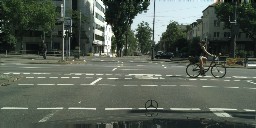}
        \includegraphics[width=0.48\linewidth]{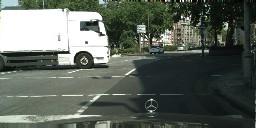}
        \includegraphics[width=0.48\linewidth]{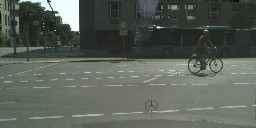}
        \includegraphics[width=0.48\linewidth]{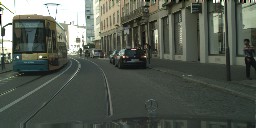}
        \includegraphics[width=0.48\linewidth]{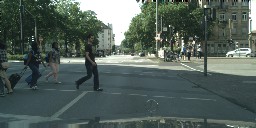}
        \includegraphics[width=0.48\linewidth]{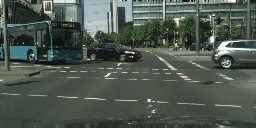}
        \includegraphics[width=0.48\linewidth]{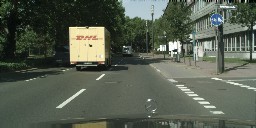}
        \includegraphics[width=0.48\linewidth]{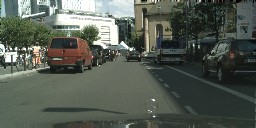}
        \includegraphics[width=0.48\linewidth]{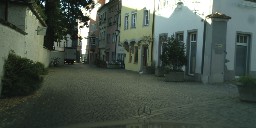}
        \includegraphics[width=0.48\linewidth]{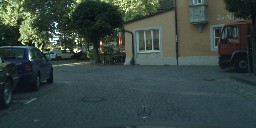}
        \includegraphics[width=0.48\linewidth]{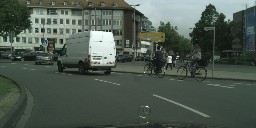}
        \includegraphics[width=0.48\linewidth]{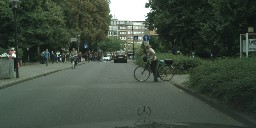}
        \includegraphics[width=0.48\linewidth]{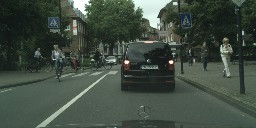}
        \includegraphics[width=0.48\linewidth]{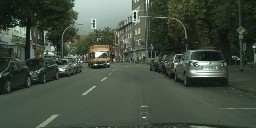}
        \includegraphics[width=0.48\linewidth]{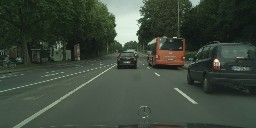}
        \includegraphics[width=0.48\linewidth]{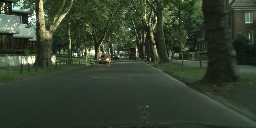}
    \end{subfigure}
    \centering
    \begin{subfigure}[t]{0.49\textwidth}
        \centering
        \caption{{Ours}}
        \includegraphics[width=0.48\linewidth]{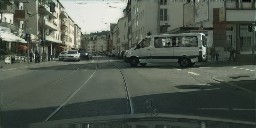}
        \includegraphics[width=0.48\linewidth]{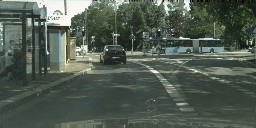}
        \includegraphics[width=0.48\linewidth]{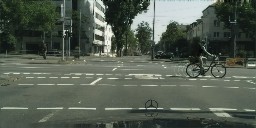}
        \includegraphics[width=0.48\linewidth]{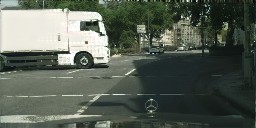}
        \includegraphics[width=0.48\linewidth]{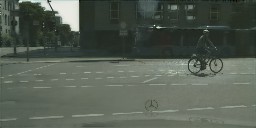}
        \includegraphics[width=0.48\linewidth]{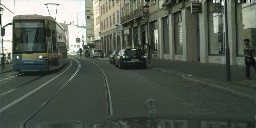}
        \includegraphics[width=0.48\linewidth]{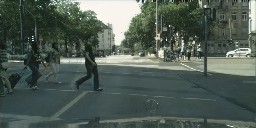}
        \includegraphics[width=0.48\linewidth]{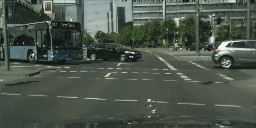}
        \includegraphics[width=0.48\linewidth]{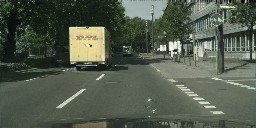}
        \includegraphics[width=0.48\linewidth]{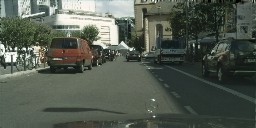}
        \includegraphics[width=0.48\linewidth]{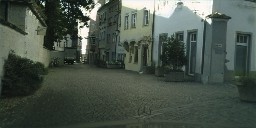}
        \includegraphics[width=0.48\linewidth]{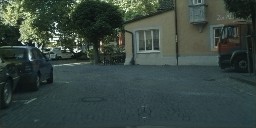}
        \includegraphics[width=0.48\linewidth]{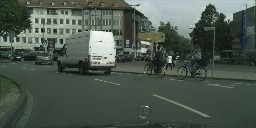}
        \includegraphics[width=0.48\linewidth]{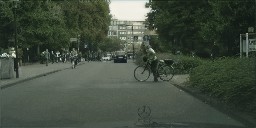}
        \includegraphics[width=0.48\linewidth]{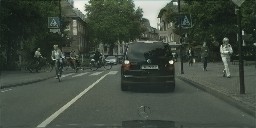}
        \includegraphics[width=0.48\linewidth]{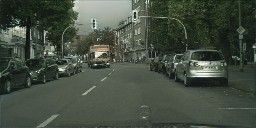}
        \includegraphics[width=0.48\linewidth]{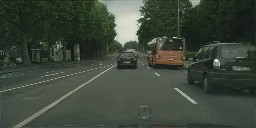}
        \includegraphics[width=0.48\linewidth]{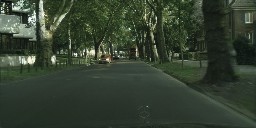}
    \end{subfigure}
    \caption{Output of the proposed algorithm obtained on the \textit{Cityscapes} dataset~\cite{cordts2016cityscapes}.}
    \label{fig:samples_cs}
\end{figure*}

In the following, we report sample visual results of the proposed model and of the competitors~\cite{munda2018real,scheerlinck2018continuous}.
In particular, in Figure~\ref{fig:gen_ddd} and in Figure~\ref{fig:gen_mvsec} we extend Figure $5$ of the original paper, including more synthesized images taken from event-based datasets~\cite{binas2017ddd17,zhu2018multi}.
Figure~\ref{fig:cross_ddd} and~\ref{fig:cross_mvsec} show frames obtained from the cross-modality test (see Sect.~$5.3$ in the paper). 
In Figure~\ref{fig:samples_cs} and~\ref{fig:samples_kitti} images generated from \textit{Kitti}~\cite{Geiger2013IJRR} and \textit{Cityscapes}~\cite{cordts2016cityscapes} datasets are reported.\\
Finally, from Figure~\ref{fig:ss_kitti} to Figure~\ref{fig:od_cs} we extend Figure $6$, adding ground truth semantic segmentation maps, together with semantic maps computed both on original and generated frames. The same procedure has been adopted with images reporting the object detection samples, represented through superimposed colored bounding boxes.



\begin{figure*}[th]
\captionsetup[subfigure]{labelformat=empty}
\centering
\resizebox{0.9\textwidth}{!}{
    \centering
    \begin{subfigure}[t]{0.19\textwidth}
        \centering
        \caption{{\small Original}}
        \includegraphics[width=0.98\linewidth]{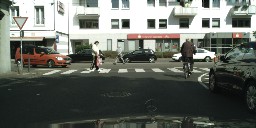}
        \includegraphics[width=0.98\linewidth]{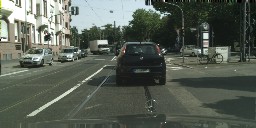}
        \includegraphics[width=0.98\linewidth]{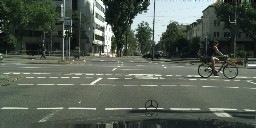}
        \includegraphics[width=0.98\linewidth]{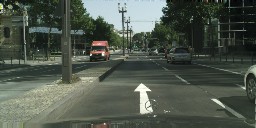}
    \end{subfigure}
    \centering
    \begin{subfigure}[t]{0.19\textwidth}
        \centering
        \caption{{\small Ground Truth}}
        \includegraphics[width=0.98\linewidth]{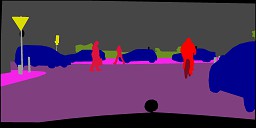}
        \includegraphics[width=0.98\linewidth]{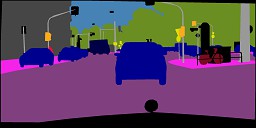}
        \includegraphics[width=0.98\linewidth]{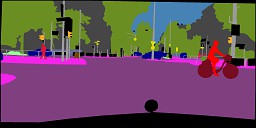}
        \includegraphics[width=0.98\linewidth]{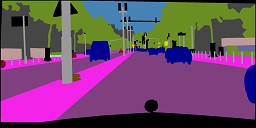}
    \end{subfigure}
    \centering
    \begin{subfigure}[t]{0.19\textwidth}
        \centering
        \caption{{\small SS on Original}}
        \includegraphics[width=0.98\linewidth]{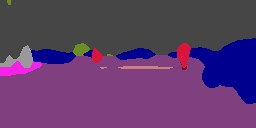}
        \includegraphics[width=0.98\linewidth]{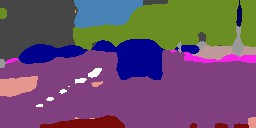}
        \includegraphics[width=0.98\linewidth]{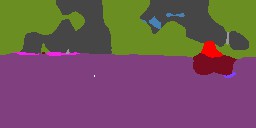}
        \includegraphics[width=0.98\linewidth]{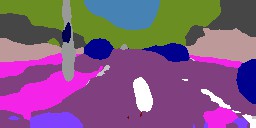}
    \end{subfigure}
    \begin{subfigure}[t]{0.19\textwidth}
        \centering
        \caption{{\small SS on Ours}}
        \includegraphics[width=0.98\linewidth]{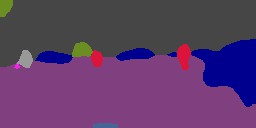}
        \includegraphics[width=0.98\linewidth]{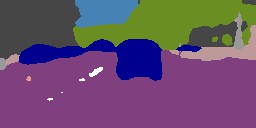}
        \includegraphics[width=0.98\linewidth]{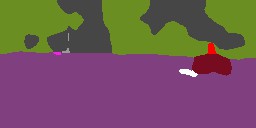}
        \includegraphics[width=0.98\linewidth]{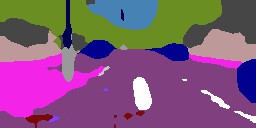}
    \end{subfigure}
}
    \caption{Semantic Segmentation applied on \textit{Cityscapes} dataset~\cite{cordts2016cityscapes}.
    From left, we report the original RGB frame, the ground truth of the semantic segmentation and then the segmentation computed on original and synthesised frames.}
    \label{fig:ss_cs}
\end{figure*}


\begin{figure*}[th]
\captionsetup[subfigure]{labelformat=empty}
\centering
\resizebox{0.7\textwidth}{!}{
    \centering
    \begin{subfigure}[t]{0.19\textwidth}
        \centering
        \caption{{\small Original}}
        \includegraphics[width=0.98\linewidth]{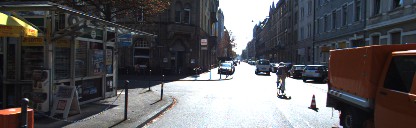}
        \includegraphics[width=0.98\linewidth]{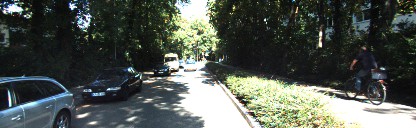}
        \includegraphics[width=0.98\linewidth]{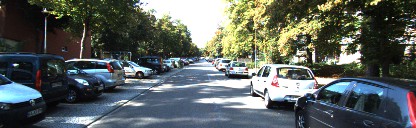}
        \includegraphics[width=0.98\linewidth]{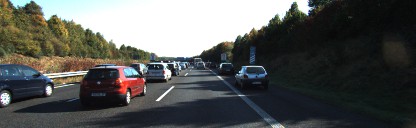}
        \includegraphics[width=0.98\linewidth]{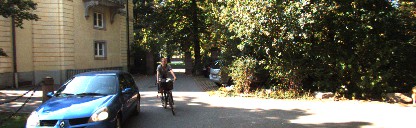}
    \end{subfigure}
    \centering
    \begin{subfigure}[t]{0.19\textwidth}
        \centering
        \caption{{\small OD on original}}
        \includegraphics[width=0.98\linewidth]{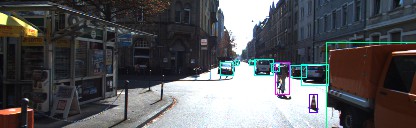}
        \includegraphics[width=0.98\linewidth]{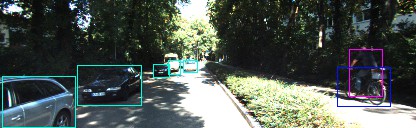}
        \includegraphics[width=0.98\linewidth]{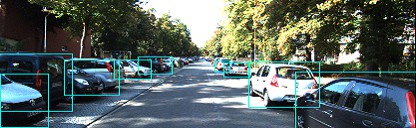}
        \includegraphics[width=0.98\linewidth]{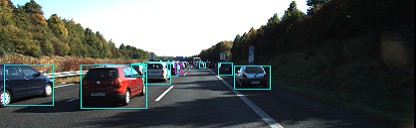}
        \includegraphics[width=0.98\linewidth]{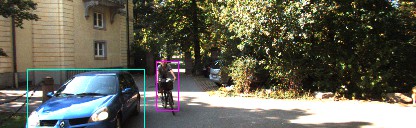}
    \end{subfigure}
    \centering
    \begin{subfigure}[t]{0.19\textwidth}
        \centering
        \caption{{\small OD on Ours}}
        \includegraphics[width=0.98\linewidth]{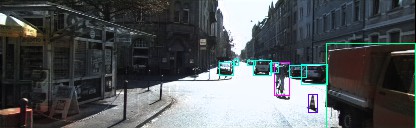}
        \includegraphics[width=0.98\linewidth]{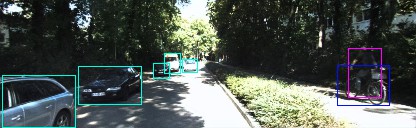}
        \includegraphics[width=0.98\linewidth]{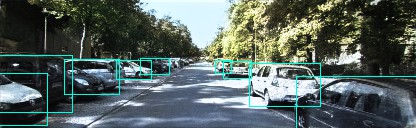}
        \includegraphics[width=0.98\linewidth]{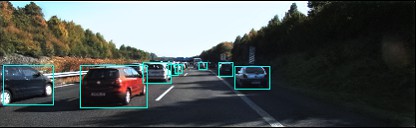}
        \includegraphics[width=0.98\linewidth]{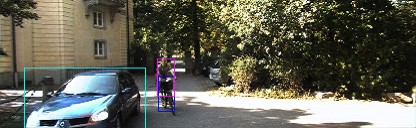}
    \end{subfigure}
}
    \caption{Object Detection on \textit{Kitti} dataset~\cite{Geiger2013IJRR}.}
    \label{fig:od_kitti}
\end{figure*}

\end{document}